\documentclass[conference]{IEEEtran}
\IEEEoverridecommandlockouts
\usepackage{cite}
\usepackage{amsmath,amssymb,amsfonts}
\usepackage{algorithmic}
\usepackage{graphicx}
\usepackage{textcomp}
\usepackage{xcolor}
\usepackage{url}
\usepackage{booktabs}
\usepackage{subfigure}

\def\BibTeX{{\rm B\kern-.05em{\sc i\kern-.025em b}\kern-.08em
    T\kern-.1667em\lower.7ex\hbox{E}\kern-.125emX}}
    
\begin{document}

\title{Drone Stereo Vision for Radiata Pine Branch Detection and Distance Measurement: Utilizing Deep Learning and YOLO Integration\\}

\author{\IEEEauthorblockN{Yida Lin, Bing Xue, Mengjie Zhang}
\IEEEauthorblockA{
\small
\textit{Centre for Data Science and Artificial Intelligence} \\
\textit{Victoria University of Wellington, Wellington, New Zealand}\\
linyida\texttt{@}myvuw.ac.nz, 
bing.xue\texttt{@}vuw.ac.nz, 
mengjie.zhang\texttt{@}vuw.ac.nz}
\and
\IEEEauthorblockN{Sam Schofield, Richard Green}
\IEEEauthorblockA{
\small
\textit{Department of Computer Science and Software Engineering} \\
\textit{Canterbury University, Canterbury, New Zealand}\\
sam.schofield\texttt{@}canterbury.ac.nz, 
richard.green\texttt{@}canterbury.ac.nz}
}

% Add the following code after the \maketitle command:
\IEEEpubidadjcol

\maketitle

\begin{abstract}
This research focuses on the development of a drone equipped with pruning tools and a stereo vision camera to accurately detect and measure the spatial positions of tree branches. YOLO is employed for branch segmentation, while two depth estimation approaches, monocular and stereo, are investigated. In comparison to SGBM, deep learning techniques produce more refined and accurate depth maps. In the absence of ground-truth data, a fine-tuning process using deep neural networks is applied to approximate optimal depth values. This methodology facilitates precise branch detection and distance measurement, addressing critical challenges in the automation of pruning operations. The results demonstrate notable advancements in both accuracy and efficiency, underscoring the potential of deep learning to drive innovation and enhance automation in the agricultural sector.
\end{abstract}


\section{Introduction}

Radiata pine (Pinus radiata) is a highly valuable species extensively cultivated in New Zealand due to its rapid growth and significant contribution to the forestry and timber industries \cite{mead2013sustainable}. Its wood is crucial for construction and manufacturing, making it a key economic asset for the country \cite{maclaren1993radiata}. Regular pruning is essential to maintain wood quality and promote the development of straight trunks \cite{kimberley2015modelling}; however, manual pruning is labor-intensive and poses substantial safety risks \cite{bentley2005understanding}. Consequently, there is a pressing need for automated solutions to address these challenges.

In response, we propose an autonomous drone equipped with a stereo camera and a pruning tool capable of detecting and trimming branches as thin as 10mm. By leveraging stereo vision for branch identification and distance measurement, our system aims to enhance the precision, efficiency, and cost-effectiveness of drone-assisted pruning \cite{linDroneStereoVision2024}.

To develop and evaluate this system, we already systematically collected indoor data using a ZED Mini camera\footnote{\url{https://store.stereolabs.com/products/zed-mini}}, capturing high-resolution images (1920×1080) from various strategic locations within a laboratory setting. Tree branches were photographed under diverse lighting conditions to create a robust and representative dataset, mitigating the risk of overfitting to idealized training data. This process resulted in 61 pairs of images (totaling 122) for training and an additional 10 pairs for testing.

Utilizing this dataset, we evaluated several object detection models and found that the YOLO\cite{reis2023real}\cite{wang2024yolov9} architecture outperformed Mask R-CNN\cite{he2017mask} in accurately detecting tree branches. Additionally, the application of Semi-Global Matching (SGBM)\cite{hirschmuller2007stereo} combined with Weighted Least Squares (WLS)\cite{daubechies2010iteratively} post-processing yielded depth maps of reasonable clarity. However, persistent pixel mismatches around tree branches continued to limit the precision of depth estimation.

Given the absence of ground-truth depth data and the limitations of depth maps produced by SGBM with WLS post-processing, the present research aims to fine-tune deep neural networks to generate depth maps that closely approximate ground-truth data. By enhancing depth estimation accuracy, these improved depth maps can be integrated into drone systems to enable precise evaluation of tree branch depth information. This advancement is expected to significantly improve drone navigation and pruning capabilities, thereby contributing to more efficient and safer forestry management operations.

\begin{figure}[htbp]
\centering
\includegraphics[width=8cm]{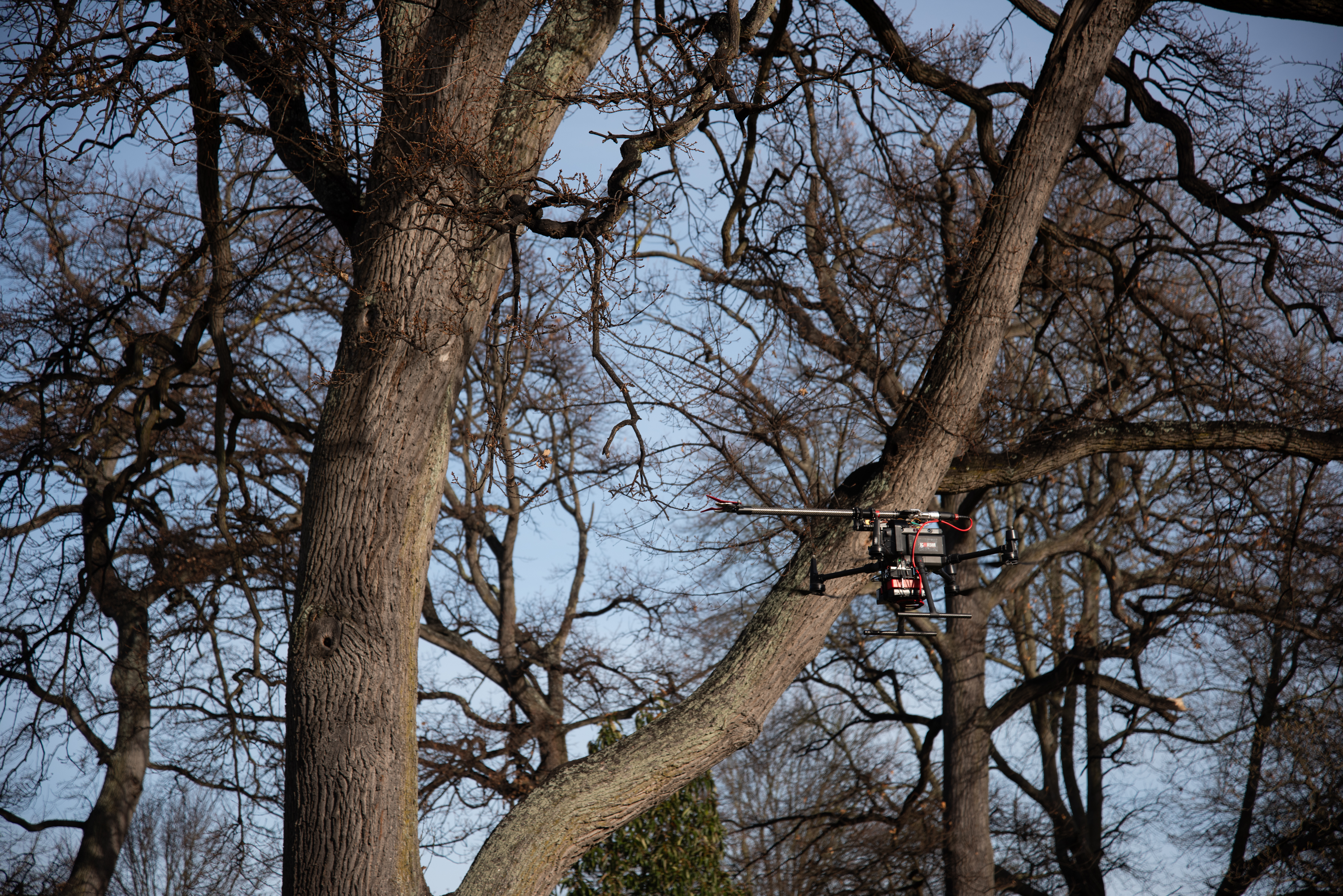}
\caption{The drone, equipped with a ZED mini camera for stereo vision and a pruning tool autonomously detects and prunes branches of radiata pine. The ZED mini camera enables the drone to accurately identify the branches, while the pruning tool precisely prunes them.}
\label{drone}
\end{figure}

\section{Related Work}

Depth map generation is a fundamental component of computer vision, enabling the reconstruction of a scene's three-dimensional structure from visual data \cite{laga2020survey}. In our drone application, which utilizes a stereo camera system such as the ZED Mini, depth maps are computed from two distinct viewpoints. These maps are essential for the precise identification of tree branches and the accurate measurement of their distance from the drone, thereby facilitating effective pruning operations.

Traditional depth estimation methods like Block Matching (BM) \cite{li1994new,po1996novel} and Semi-Global Block Matching (SGBM) \cite{hirschmuller2007stereo} have been widely employed. BM is a local search-based technique that calculates depth values by finding the best match within a fixed window, making it suitable for real-time applications. However, it is prone to errors in regions with sparse textures or occlusions. SGBM improves upon BM by introducing a semi-global cost aggregation strategy, enhancing the accuracy and robustness of depth estimation by optimizing pixel correspondences across the entire image—a method particularly effective in handling texture-rich scenes.

With the advancement of deep learning techniques, neural network-based methods have significantly improved the accuracy and efficiency of depth estimation. Stereo matching networks such as ACVNet \cite{xu2022attention} and GWCNet \cite{guo2019group} introduce sophisticated learning mechanisms to process stereo images, generating more accurate depth maps by learning matching information from paired images through deep models. MobileStereoNet \cite{shamsafar2022mobilestereonet} optimizes the network architecture for mobile devices, achieving efficient depth prediction suitable for resource-constrained environments. PSMNet \cite{chang2018pyramid} enhances feature extraction and depth estimation through pyramid pooling and 3D convolutions, while RAFT-Stereo \cite{lipson2021raft} employs recurrent neural networks to perform pixel-level correlation analysis, further improving the accuracy of depth estimation.

Innovative approaches like Neural Radiance Fields (NeRF) \cite{tosi2023nerf} have also emerged, demonstrating the capability to reconstruct complex 3D scenes through the generation of dense depth maps using deep learning. Additionally, monocular depth estimation methods, including models like MiDaS \cite{birkl2023midas} and Depth Anything \cite{yang2024depth}, leverage single-camera input to estimate depth information. These models are suitable for applications where stereo or LiDAR setups are impractical, with MiDaS recognized for its versatility across various environments and Depth Anything extending this capability by enabling depth estimation in a wide array of scenes with improved generalization.

In this research, we focus on improving depth maps without relying on ground-truth data, diverging from traditional methods. By employing deep neural network architectures based on stereo and monocular inputs, we aim to generate accurate depth maps of tree branches. This approach enhances the effectiveness of drone-assisted pruning operations by providing precise depth information crucial for manipulation tasks.

\section{Methods}
\begin{figure}[htbp]
\centering
\includegraphics[width=8cm]{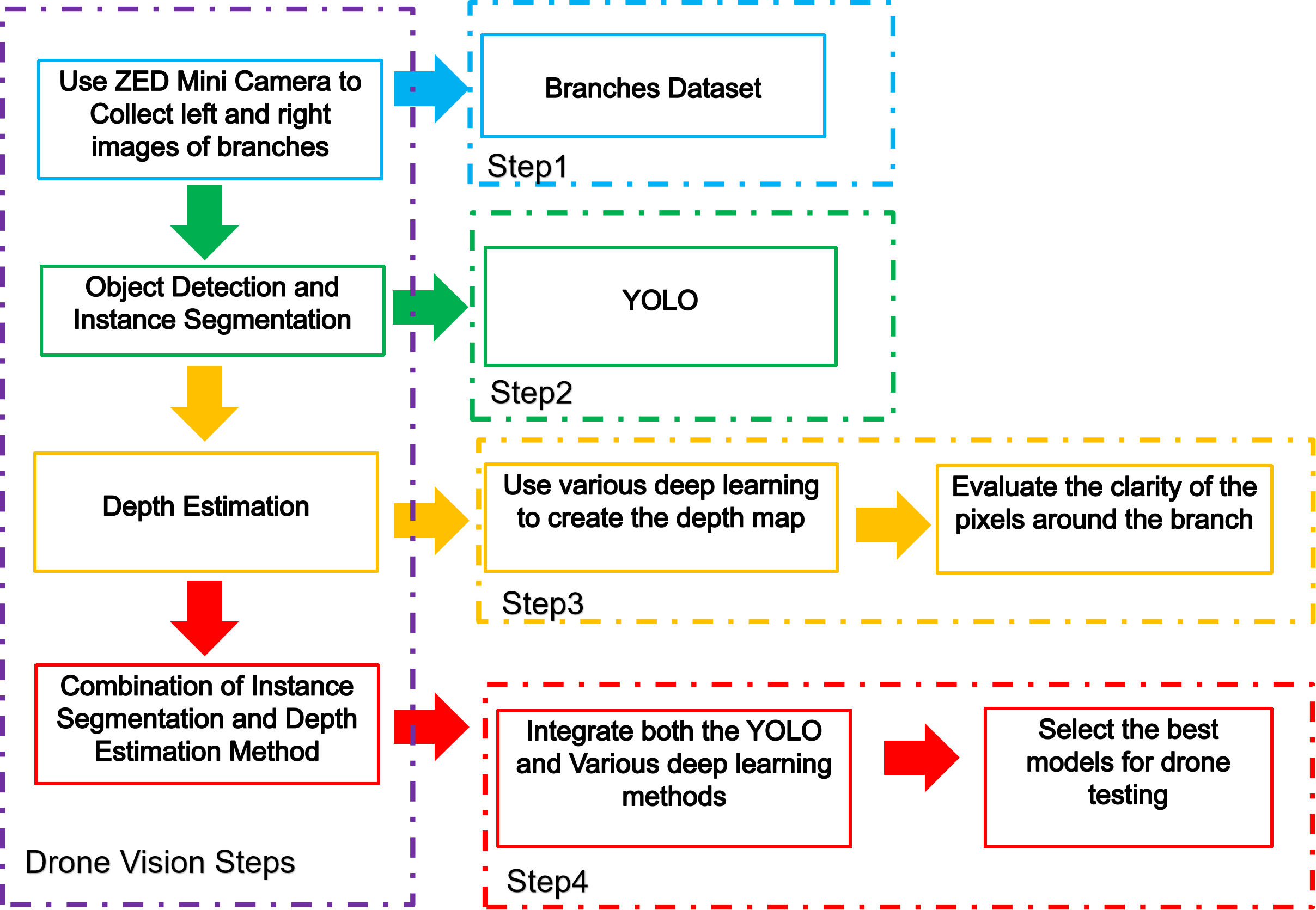}
\caption{Workflow Diagram of the Research Process}
\label{flow_chart}
\end{figure}

% here we need write more detail!
In this section, we focus on depth map generation for the small branches dataset, where YOLO is employed for both detection and segmentation. By integrating YOLO with additional deep learning techniques, we systematically process the data to achieve precise branch detection and accurate depth estimation using a stereo vision camera. The complete workflow is depicted in Fig. \ref{flow_chart}.

\subsection{Branches Data Preparation and Application of the YOLO Instance Segmentation Method}

\begin{figure}[htbp]
\centering
\includegraphics[width=8cm]{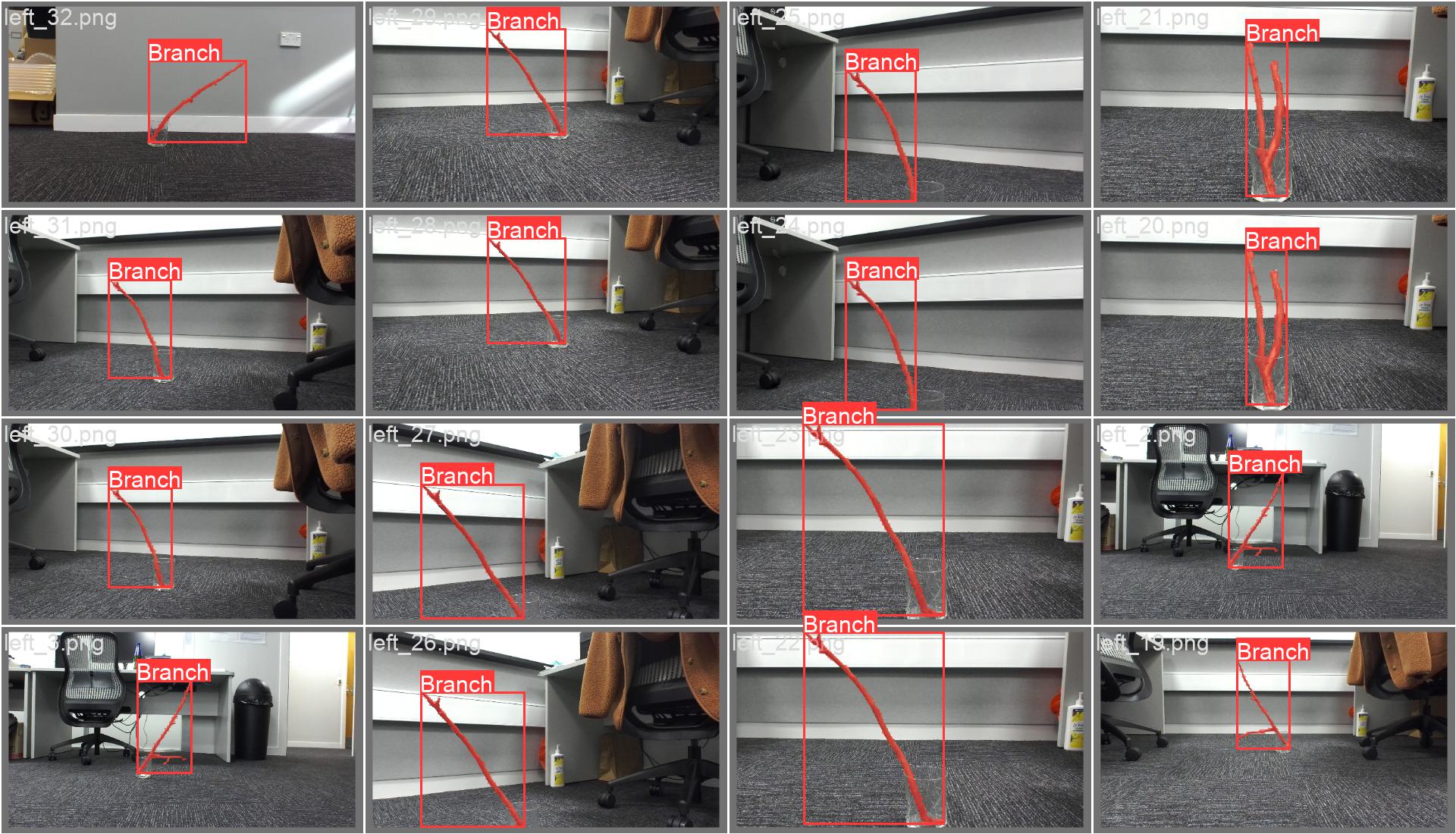}
\caption{Prediction Results Using the YOLOv8n-seg Model Trained for 100 Epochs on the Branches Dataset}
\label{YOLO_v8n}
\end{figure}

Previous research\cite{linDroneStereoVision2024} have collected and annotated indoor branch data, resulting in a relatively small dataset, and conducted comparative analyses of various Mask R-CNN backbones and different versions of YOLO, including YOLO v8 and v9. The findings demonstrate that YOLO achieves highly accurate branch segmentation, with YOLOv8n-seg exhibiting the smallest number of parameters (Params) and computational complexity (FLOPs) \cite{yolov8}, making it the most efficient model for this dataset. Accordingly, YOLOv8n-seg was selected for this research. As illustrated in Fig. \ref{YOLO_v8n}, the model accurately segments branch contours in the images.

\subsection{Deep learning Methods for Generating Depth Map}

% Monocular Deep Learning & % Stereo Deep Learning

Compared to traditional algorithms, deep learning methods for generating depth maps primarily focus on two areas: monocular depth estimation and stereo depth estimation. In the field of monocular depth estimation, two leading models are MiDaS\cite{birkl2023midas} and Depth Anything\cite{yang2024depth}. MiDaS employs transformer-based backbones such as BEiT, Swin, and SwinV2, trained on large, diverse datasets using a scale-and-shift-invariant loss function, enabling robust cross-dataset zero-shot inference. In contrast, Depth Anything leverages vast amounts of unlabeled data while retaining rich semantic priors from pre-trained models. This approach allows Depth Anything to excel in zero-shot depth estimation and serve as a robust foundation for metric depth estimation. We directly applied MiDas and Depth Anything to our own dataset to observe its prediction results.

In the domain of stereo depth estimation, widely used methods include ACVNet\cite{xu2022attention}, GwcNet\cite{guo2019group}, PSMNet\cite{chang2018pyramid}, and RAFT-Stereo\cite{lipson2021raft}. Additionally, NeRF-Supervised Deep Stereo\cite{tosi2023nerf} offers significant advantages by generating high-quality depth maps without requiring ground-truth depth data.

In our research, we selected several commonly used stereo matching datasets, including KITTI\cite{geiger2012we} and Scene Flow\cite{mayer2016large}. Each model architecture was thoroughly fine-tuned on these datasets to ensure adaptability and robustness across different scenarios.

We then examined the impact of fine-tuning on disparity map predictions using PSMNet as a representative model. Specifically, we fine-tuned the model on three datasets—KITTI2012, KITTI2015, and Scene Flow—and conducted detailed testing. The model was trained for 100 epochs on the KITTI2012 and KITTI2015 datasets, which primarily consist of stereo image pairs from real-world urban street scenes with high-quality ground-truth data. During the fine-tuning process, we observed that, despite differences in scenes and labeling across datasets, extended training on a single dataset significantly improved the model’s prediction accuracy.

Finally, we fine-tuned various models on our own dataset. Despite the absence of ground-truth data for precise quantitative evaluation, we assessed the accuracy of depth predictions around tree branches by visually inspecting the clarity of pixel depth information. The comparative experimental results are presented and discussed in Section~\ref{Comparative_Depth_map}."

\subsection{Integration of Image Instance Segmentation and Depth Map Generation}

In the preceding sections, we have extensively examined instance segmentation and depth estimation as independent tasks. However, the overarching goal is to equip the stereo camera mounted on the drone with the ability to perform both instance segmentation and depth estimation concurrently, enabling precise spatial localization of tree branches. Achieving this requires the integration of the segmentation and depth estimation models.

Previous integration approaches\cite{linDroneStereoVision2024} involved connecting all segmented points into a plane, recording their coordinates, and mapping these points to their corresponding positions on the depth map, thereby generating a distribution. This method ultimately estimated the distance between the branch and the camera. In contrast, we propose a more efficient approach, which is mathematically formalized in this research.

Our method aims to identify the coordinates of points surrounding the branch and apply the Triangulation Method algorithm to accurately determine their locations. By correlating these points with their respective positions on the depth map, which represent the distances from the points to the camera plane, we refine the results through median calculation, outlier removal, and averaging. This process yields a precise estimation of the distance between the branch and the camera plane.

We define the $P$ be the set of the predict points, and the $i$\textsuperscript{th} point is $p_i$, Assume we have $n$ points, so $P=\{p_1, p_2,..., p_n\}$, $0<i\leq n$. and $p_i=(x_i,y_i)$, Where $x_i$ is the horizontal coordinate value of point $p_i$ in the depth map, and $y_i$ is the Vertical coordinate value of point $p_i$ in its depth map.

we through the 3 points can get it's centroid point, and the centroid mostly should in the branch, let the centroid points set be the $P'$, $P'= \{p'_1, p'_2,...,p'_k\}$, $0<k\leq \frac{n}{3}$, $p'_g = (x'_g, y'_g)$. Where $x_g$ is the horizontal coordinate value of centroid $p_g$ in the depth map, and $y_g$ is the Vertical  coordinate value of centroid $p_g$ in its depth map. Let each centroid increase by \( m \) values, and the \( j\textsuperscript{th} \) point added to centroid \( p_i \) is \( p_{i,j} \). Let the expanded points set be \( P'' \), where \( P'' = \{P'_1, P'_2, \ldots, P'_n\} \), with \( P''_i = \{p_{i,1}, p_{i,2}, \ldots, p_{i,m}\} \), and \( p_{i,j} = (x''_{i,j}, y''_{i,j}) \). We then define the total points set as \( P''' = P'' + P' \). Finally, use MAD (Median Absolute Deviation) to remove values where the error is too large or too small.

\begin{equation}
    MAD = median(\lvert P''' - median(P''')\lvert)
\end{equation}
if any point in $P'''$ is in the $median(P''')\pm k \times MAD$, whose will be left and then compose $\bar{P}$, and we calulate the the mean to make sure the final distance between the branche and caerma,
\begin{equation}
    distance = mean(\bar{P})
\end{equation}
So, we get the finall distance between the camerea and branches. And the result is shown in Fig.\ref{circle_image}. Compose all of the red dots in the picture are $\bar{P}$. Then, each point is mapped to its corresponding position on the depth map, yielding a set of corresponding depth values. After applying MAD (Median Absolute Deviation) and averaging, the final distance between the tree branch and the camera is obtained.

\section{Results and Analysis}
% here we need to say something .
In this section, we present the results of monocular depth estimation for tree branches at distances of 1m, 1.5m, and 2m. Additionally, we provide a case study of PSMNet fine-tuning across different datasets, followed by the outcomes of fine-tuning various deep learning models. A comparative analysis between NeRF and SGBM at the specified distances of 1m, 1.5m, and 2m is also included. Furthermore, we present the results obtained using our Integration Centroid Point Calculation method.

\begin{figure}[htbp]
\centering
\subfigure[Predicted Depth Map of Tree Branches at a 1m Distance from the Camera Using the MiDaS Model]{\includegraphics[width=0.23\textwidth]{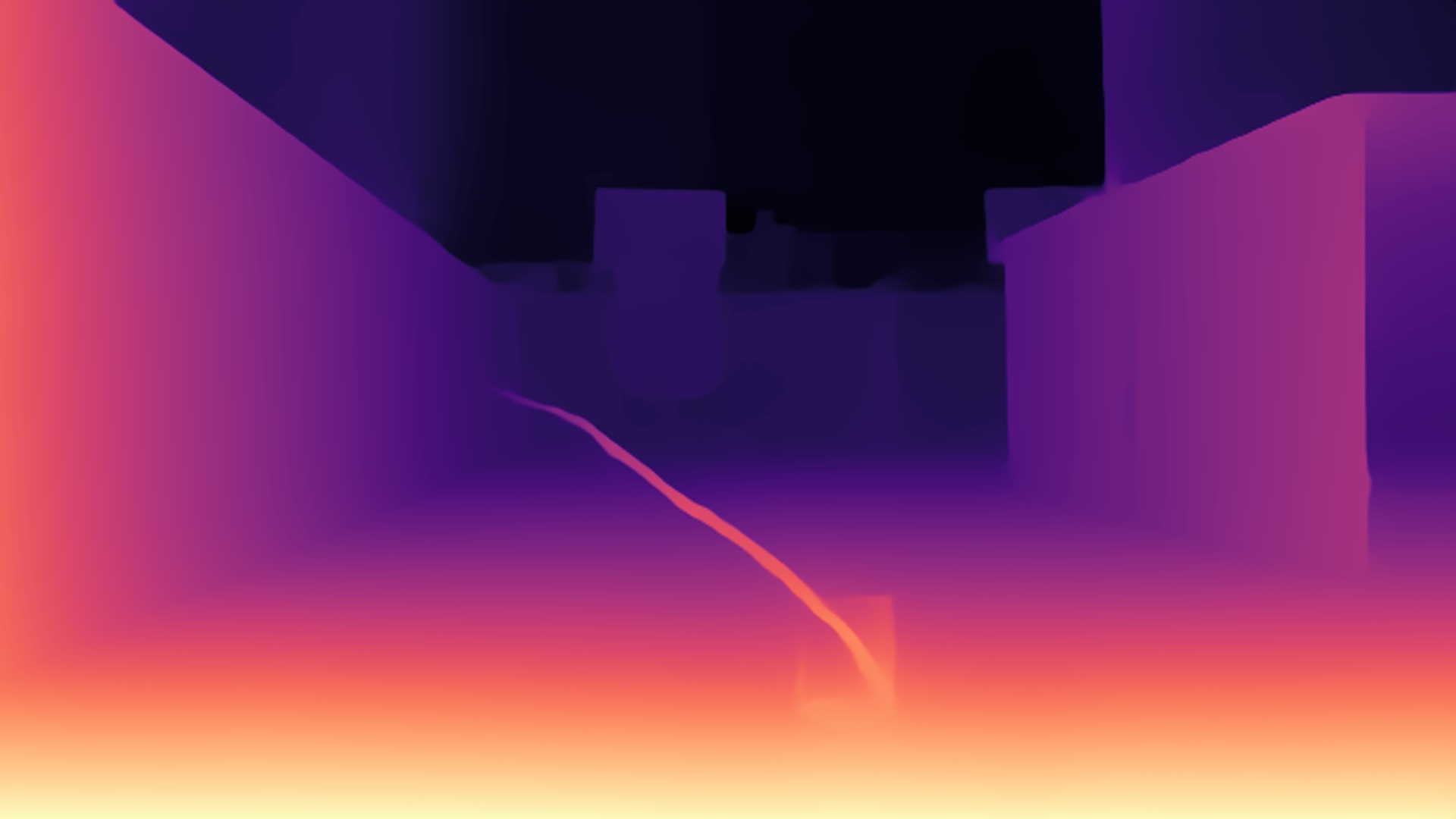}}
\subfigure[Predicted Depth Map of Tree Branches at a 1m Distance from the Camera Using the Depth Anything Model]{\includegraphics[width=0.23\textwidth]{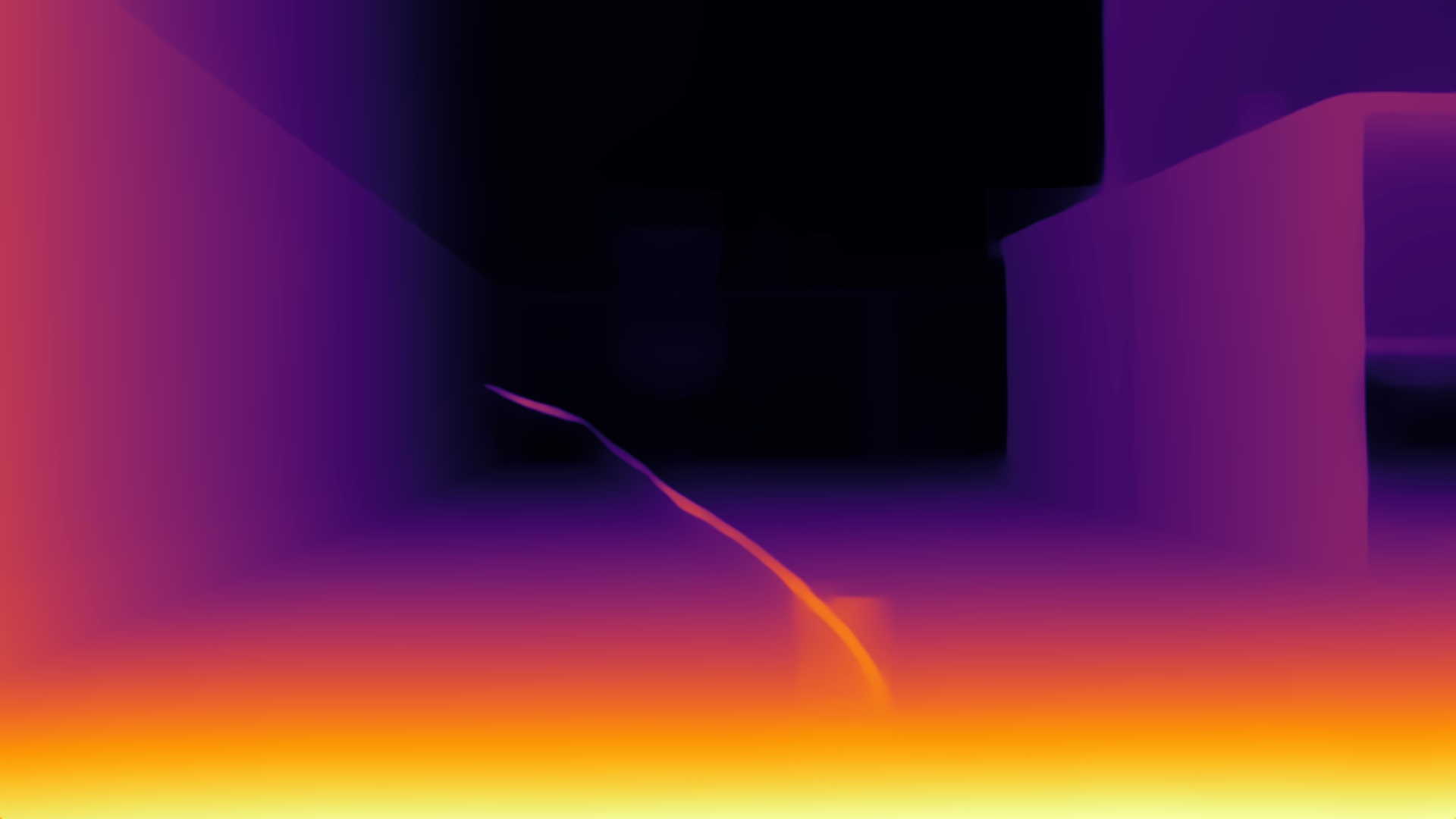}}
\subfigure[Predicted Depth Map of Tree Branches at a 1.5m Distance from the Camera Using the MiDaS Model]{\includegraphics[width=0.23\textwidth]{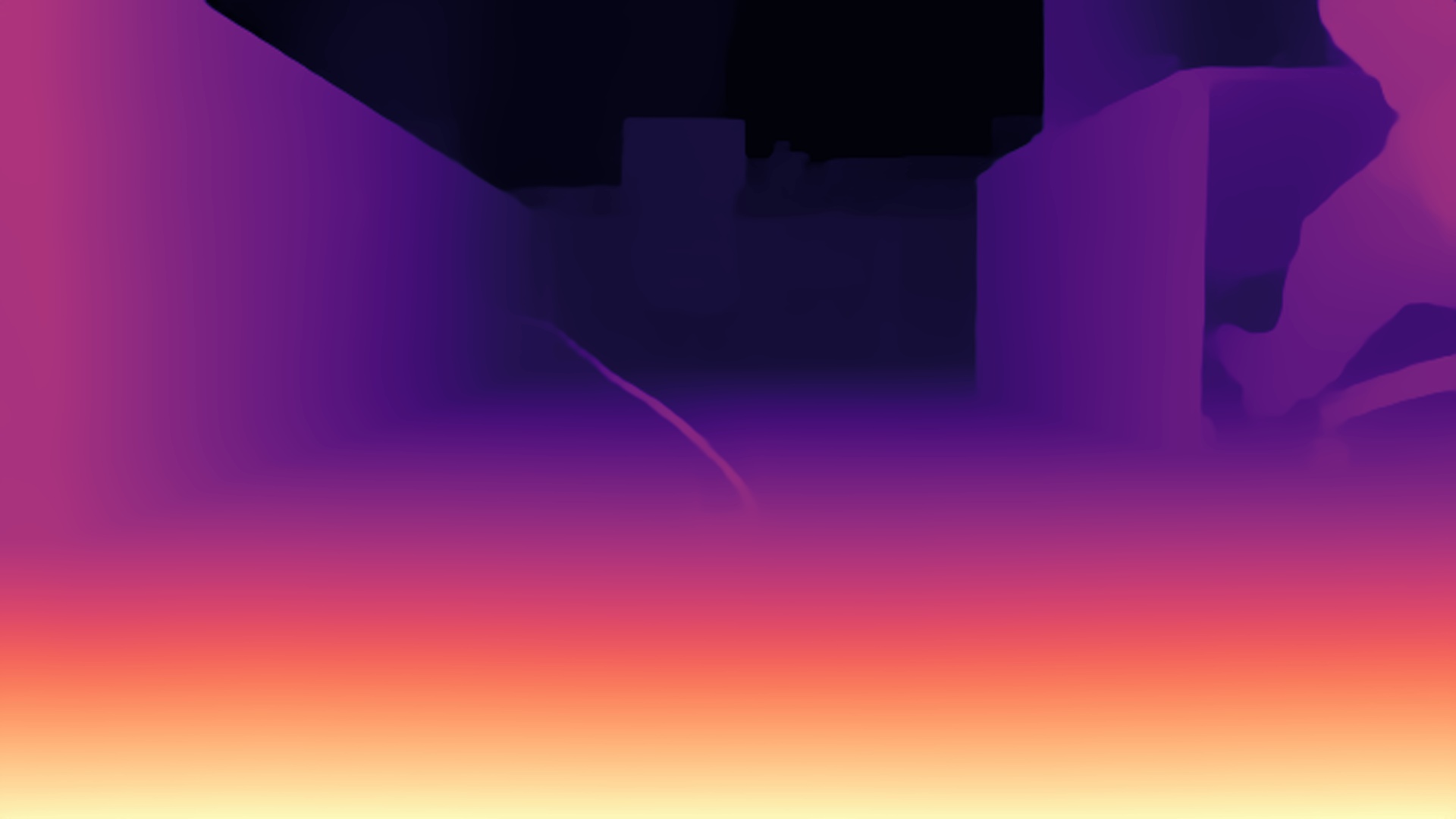}}
\subfigure[Predicted Depth Map of Tree Branches at a 1.5m Distance from the Camera Using the Depth Anything Model]{\includegraphics[width=0.23\textwidth]{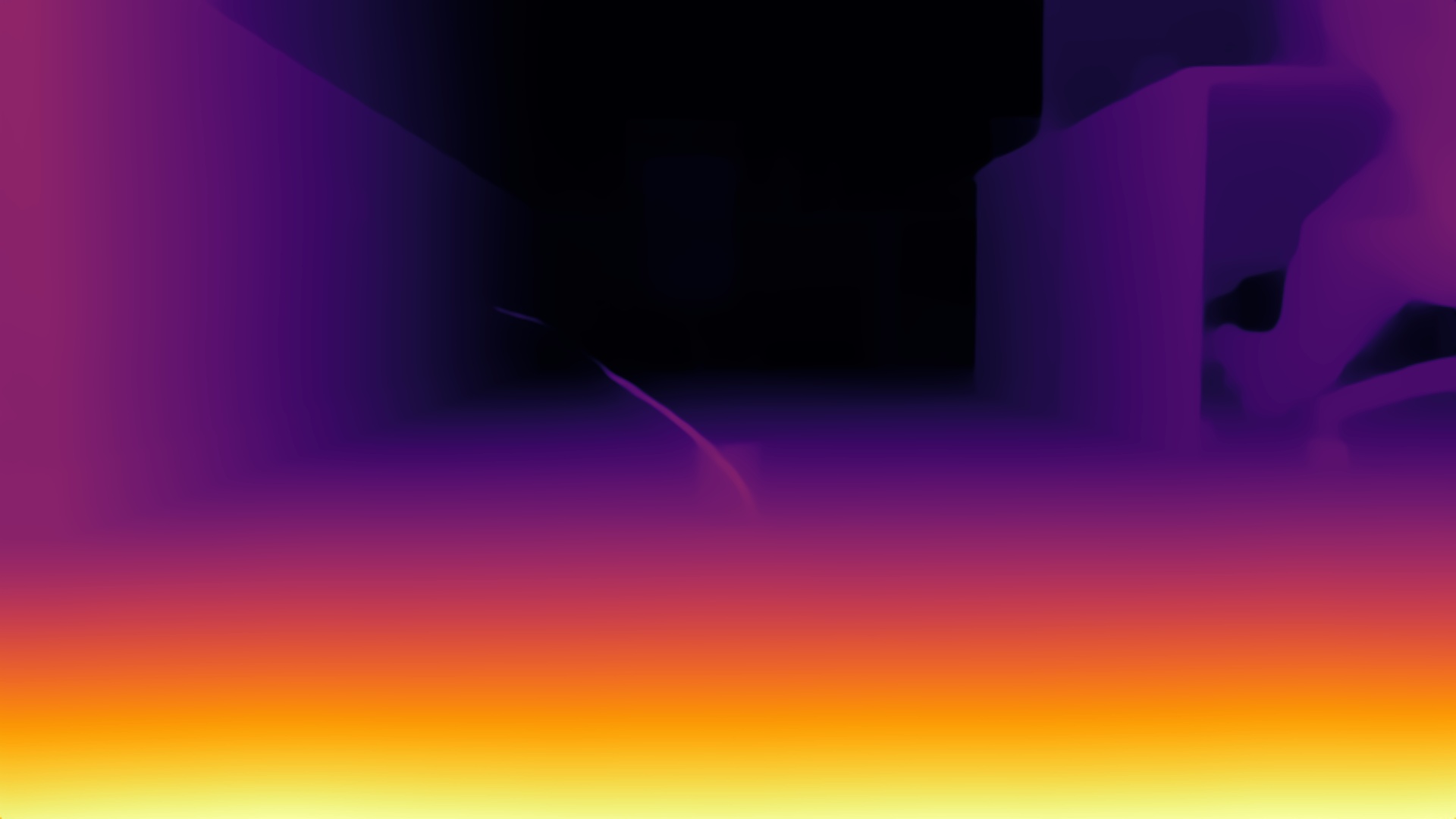}}
\vspace{\baselineskip}
\subfigure[Predicted Depth Map of Tree Branches at a 2m Distance from the Camera Using the MiDaS Model]{\includegraphics[width=0.23\textwidth]{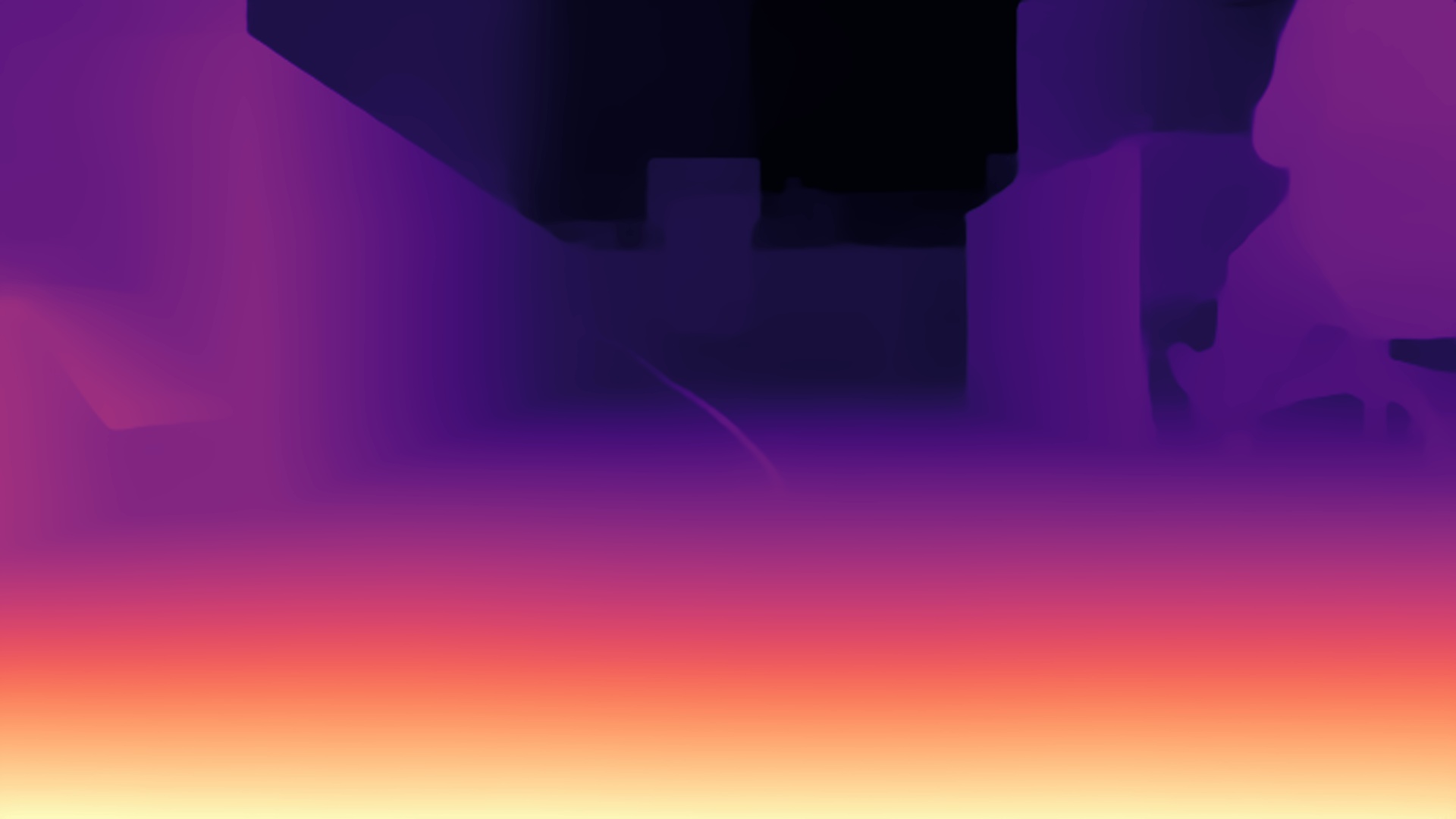}}
\subfigure[Predicted Depth Map of Tree Branches at a 2m Distance from the Camera Using the Depth Anything Model]{\includegraphics[width=0.23\textwidth]{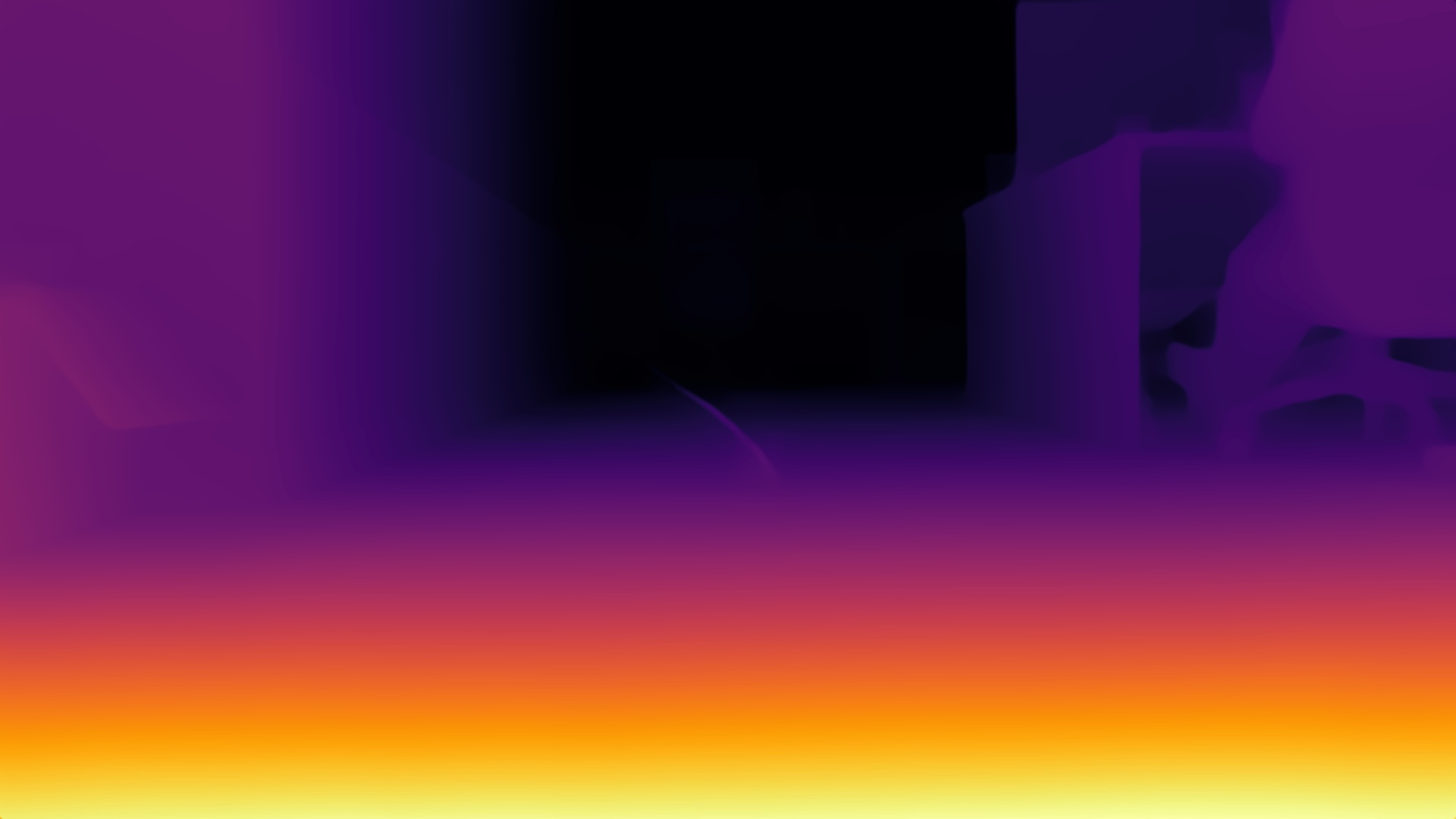}}
\caption{Comparison of Predicted Depth Maps Generated by MiDaS and Depth Anything Models at Branch Distances of 1m, 1.5m, and 2m from the Camera}
\label{Monocular}
\end{figure}

\subsection{Comparative research of Various Deep Learning Approaches for Depth Map Generation}
\label{Comparative_Depth_map}

we positioned the branches at distances of 1m, 1.5m, and 2m from the camera for testing using a monocular camera. The results are illustrated in Fig. \ref{Monocular}. At a distance of approximately 2m, the image quality deteriorates noticeably, becoming significantly blurred.

\begin{figure}[htbp]
\centering
\subfigure[original left image]{\includegraphics[width=0.23\textwidth]{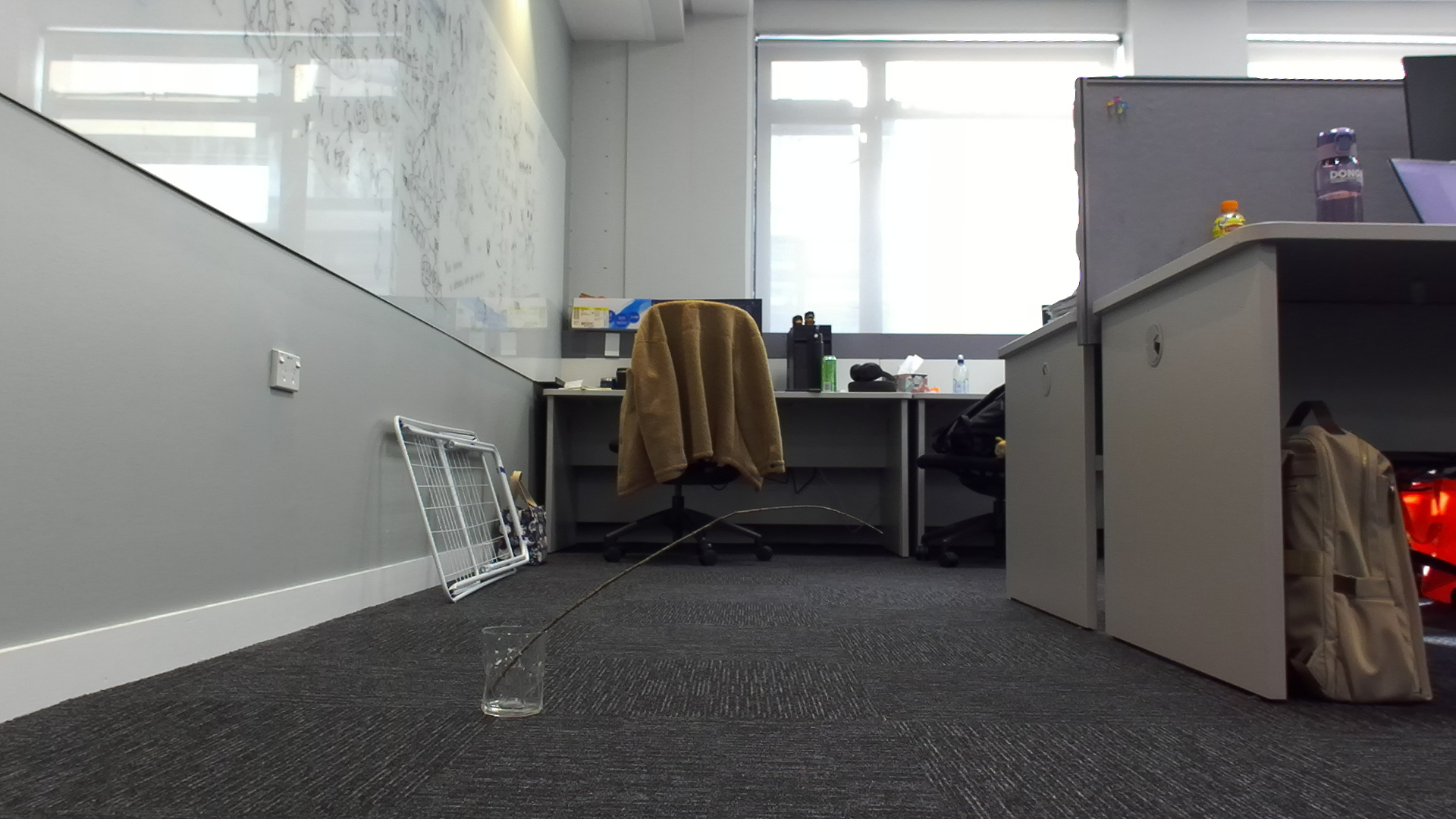}}
\subfigure[Depth Map Generation via PSMNet Fine-Tuning Using SceneFlow Pretrained Model]{\includegraphics[width=0.23\textwidth]{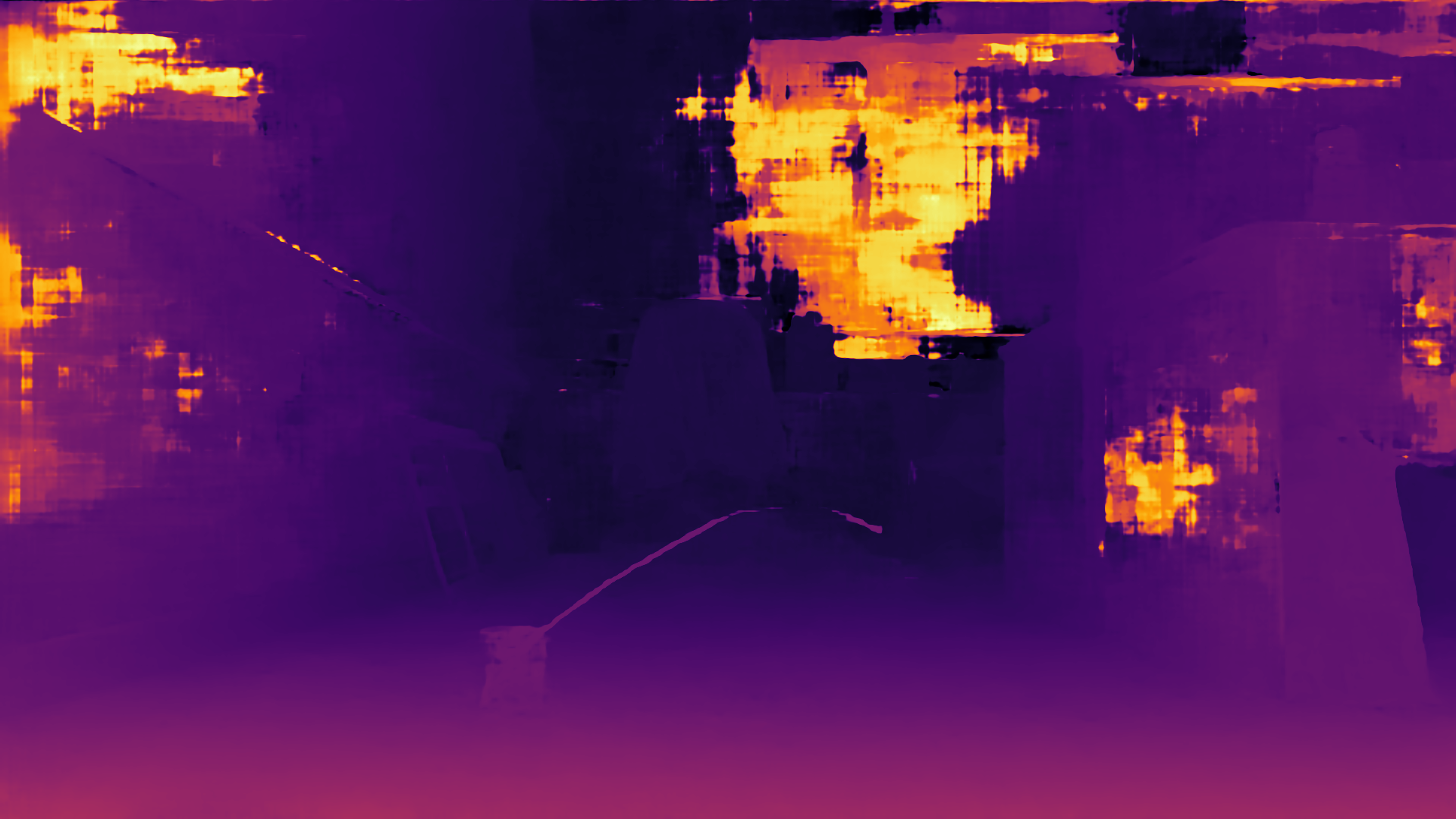}}
\subfigure[Depth Map Generation via PSMNet fine-tuning Using KITTI 2012 pretrained model]{\includegraphics[width=0.23\textwidth]{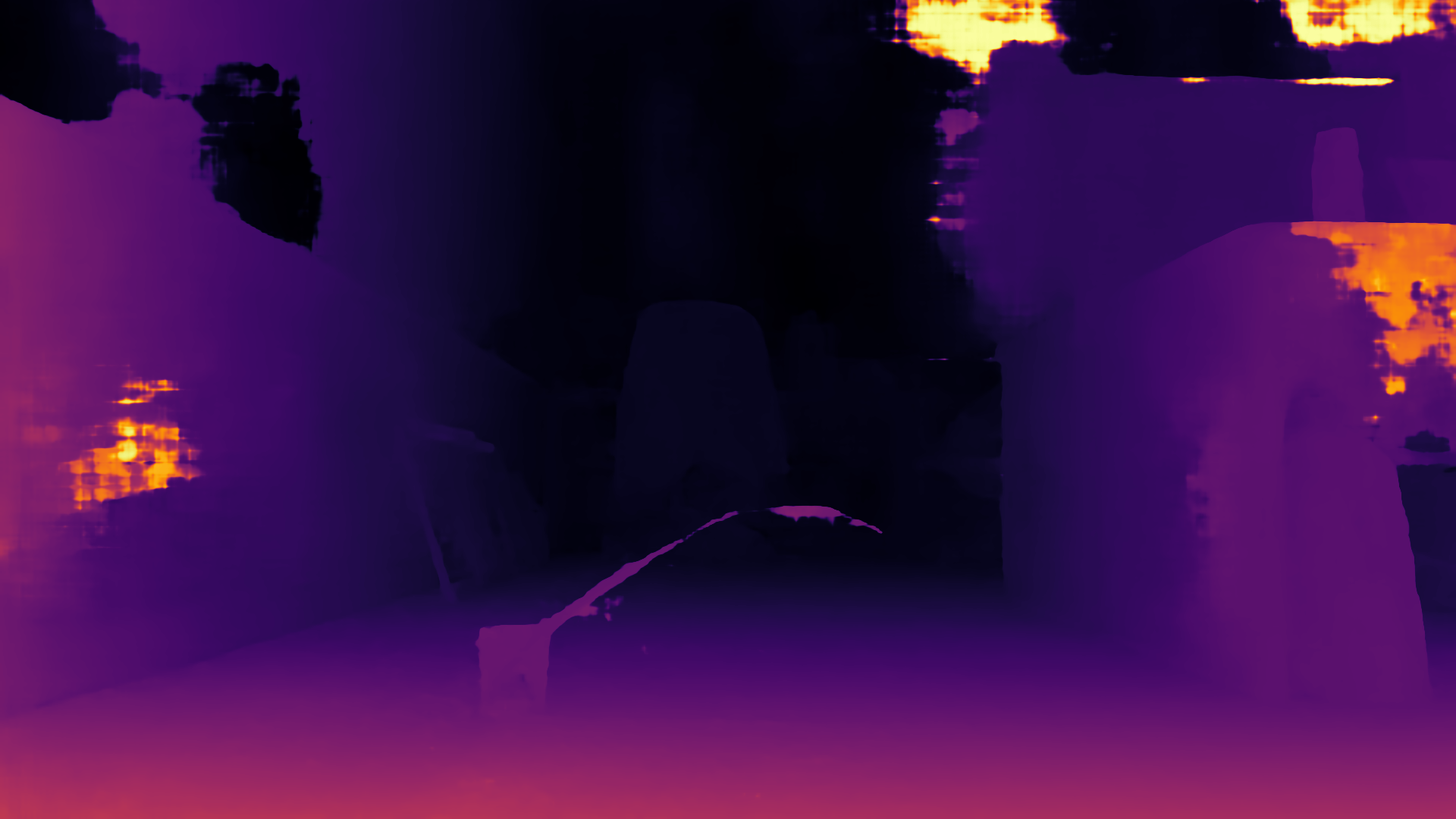}}
\subfigure[Depth Map Generation via PSMNet Fine-Tuning Using 100-Epoch KITTI 2012 Trained Model]{\includegraphics[width=0.23\textwidth]{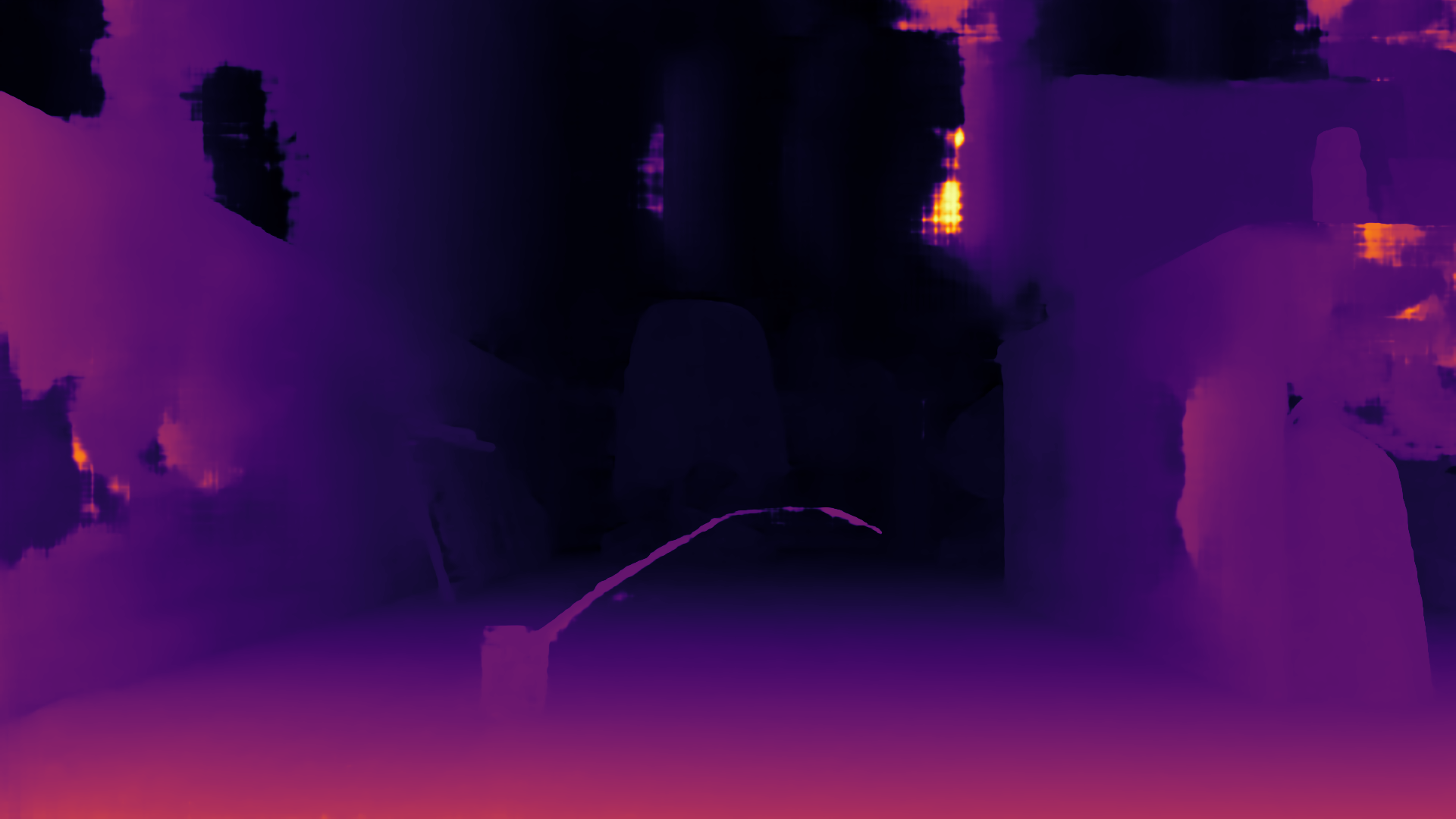}}
\subfigure[Depth Map Generation via PSMNet fine-tuning Using KITTI 2015 pretrained model]{\includegraphics[width=0.23\textwidth]{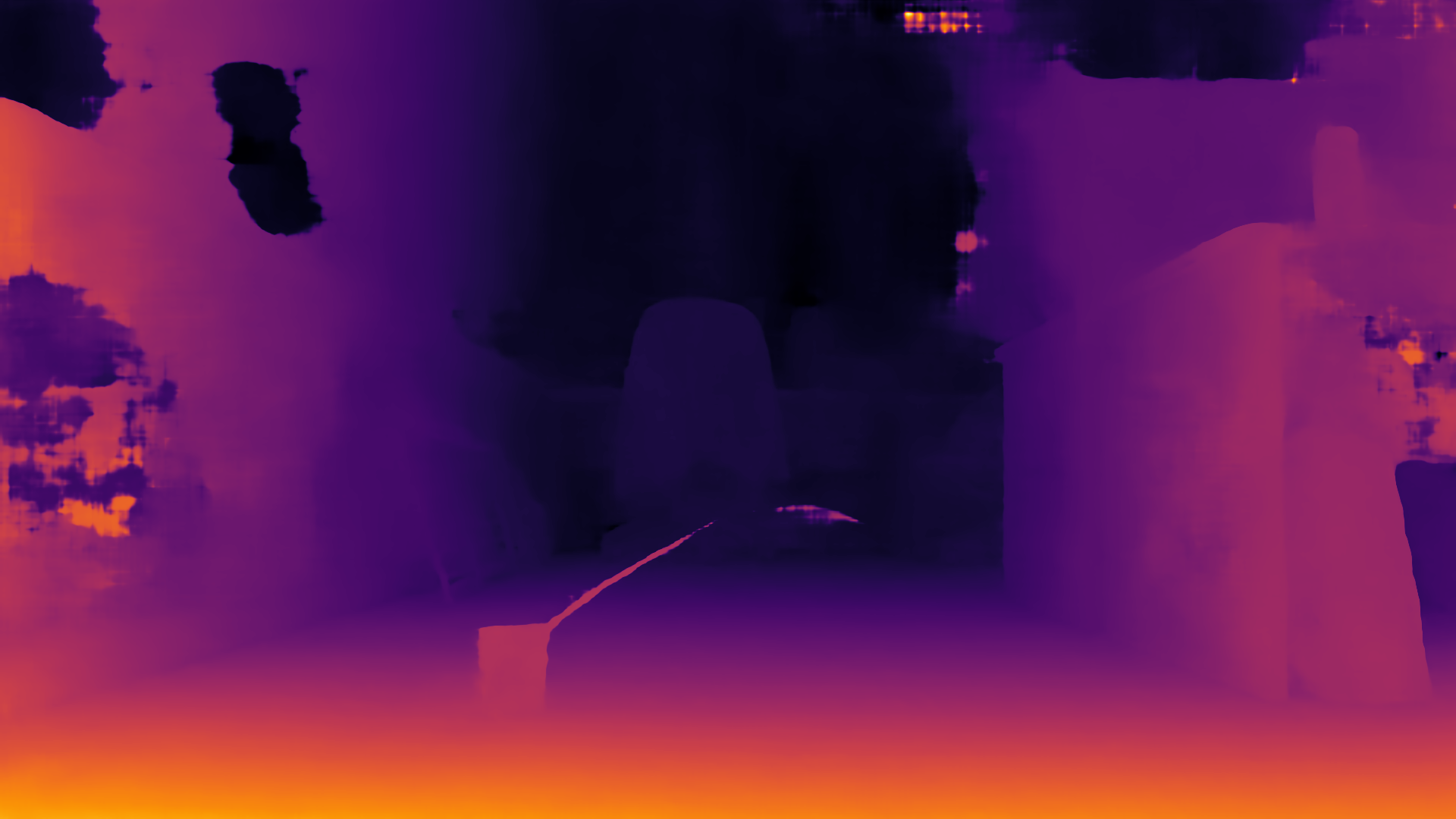}}
\subfigure[Depth Map Generation via PSMNet Fine-Tuning Using 100-Epoch KITTI 2015 Trained Model]{\includegraphics[width=0.23\textwidth]{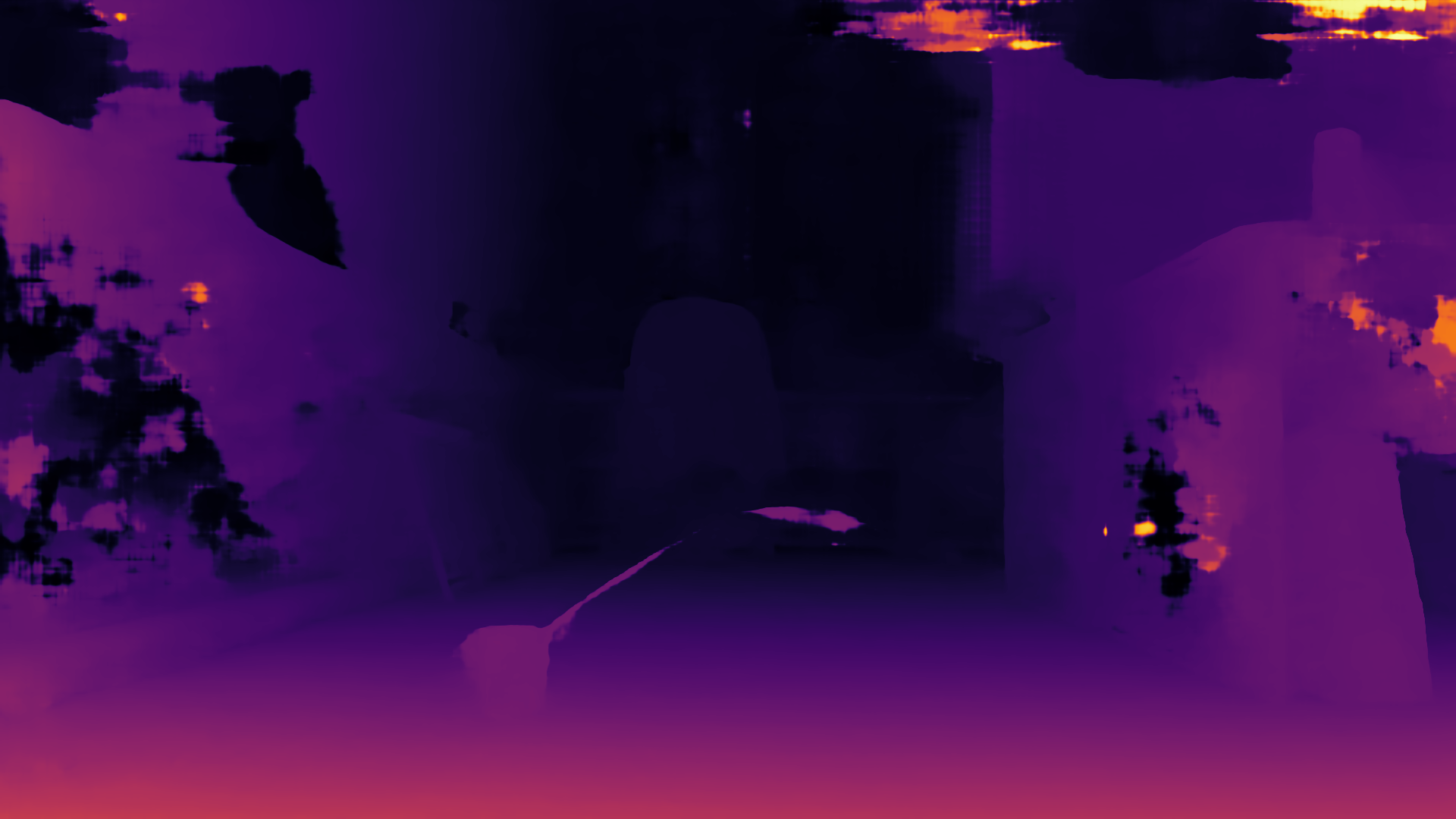}}
\caption{Comparison of PSMNet Fine-tuning Results Across Different Pretrained Models and Training Epochs}
\label{fine-tuning}
\end{figure}

% fine-tuning
And we evaluated PSMNet, a stereo depth neural network, using various pretrained models and further retraining for 100 epochs on the KITTI2012 and KITTI2015 datasets. As demonstrated in Fig. \ref{fine-tuning}, the results indicate that additional training on the KITTI datasets did not lead to improved performance when fine-tuned on our branches dataset. In fact, the performance in depth estimation deteriorated.

\begin{figure}[htbp]
\centering
\subfigure[PSMNet]{\includegraphics[width=0.23\textwidth]{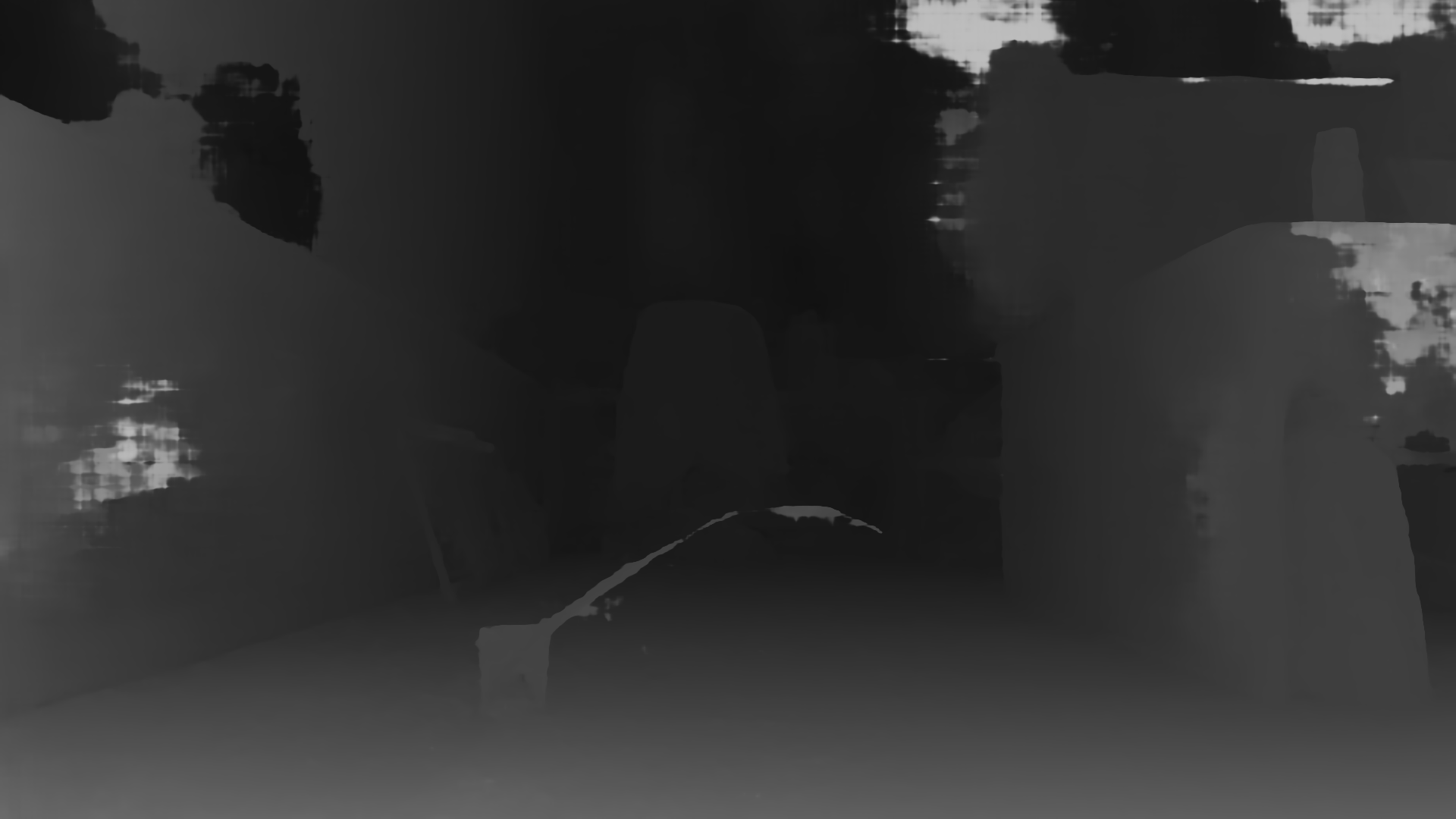}}
\subfigure[ACVNet]{\includegraphics[width=0.23\textwidth]{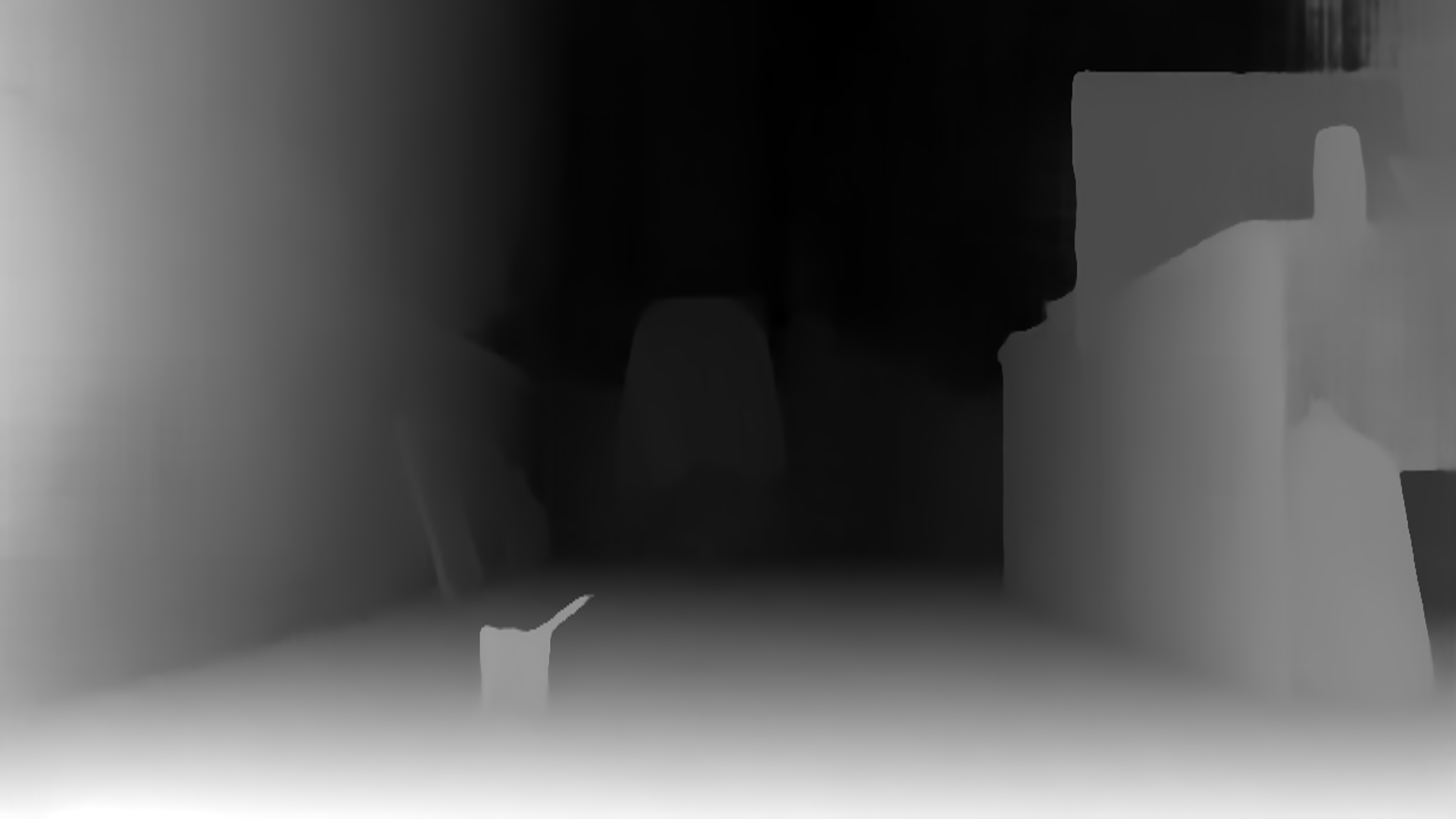}}
\subfigure[GWcNet]{\includegraphics[width=0.23\textwidth]{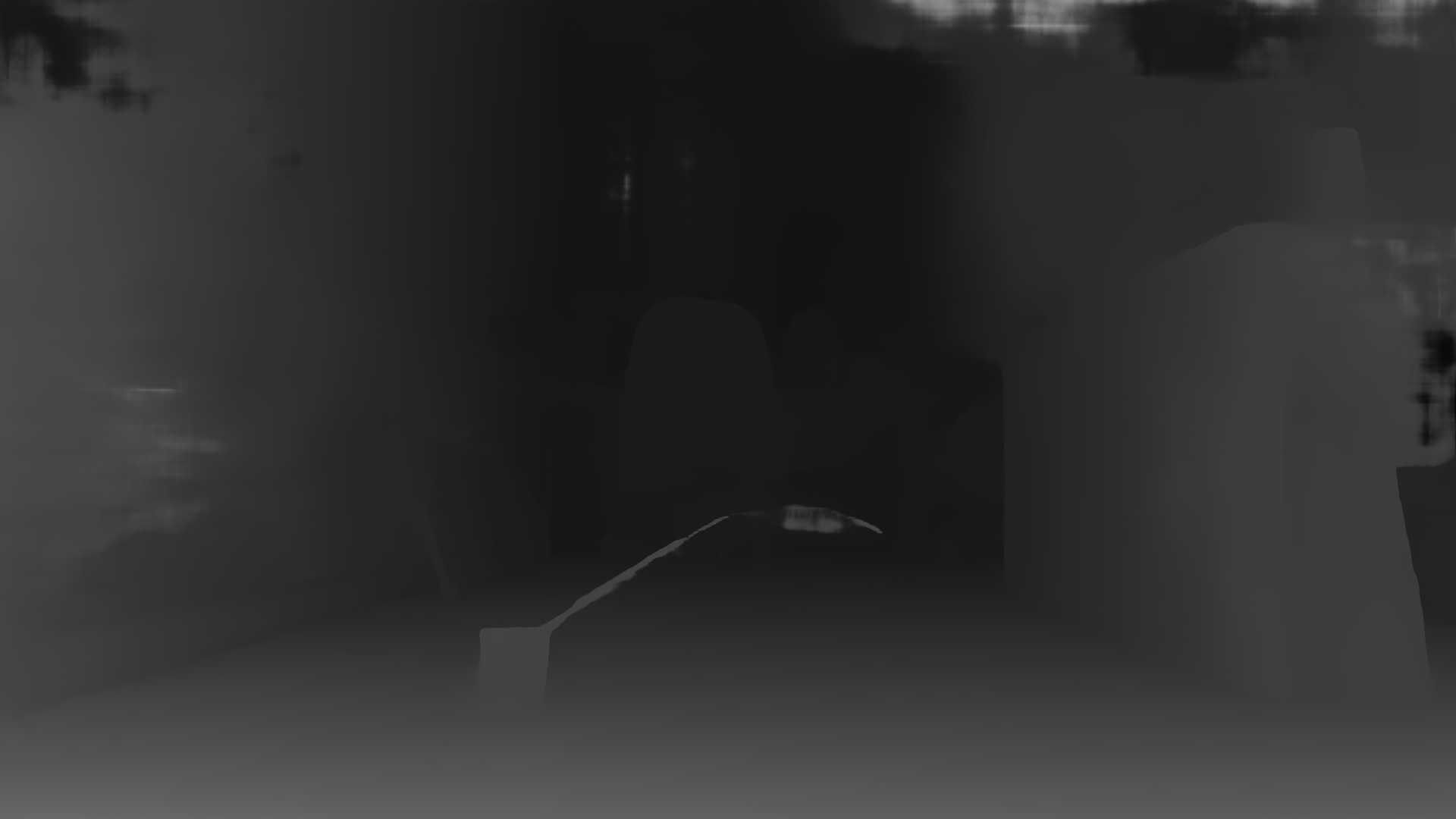}}
\subfigure[mobilesteroNet]{\includegraphics[width=0.23\textwidth]{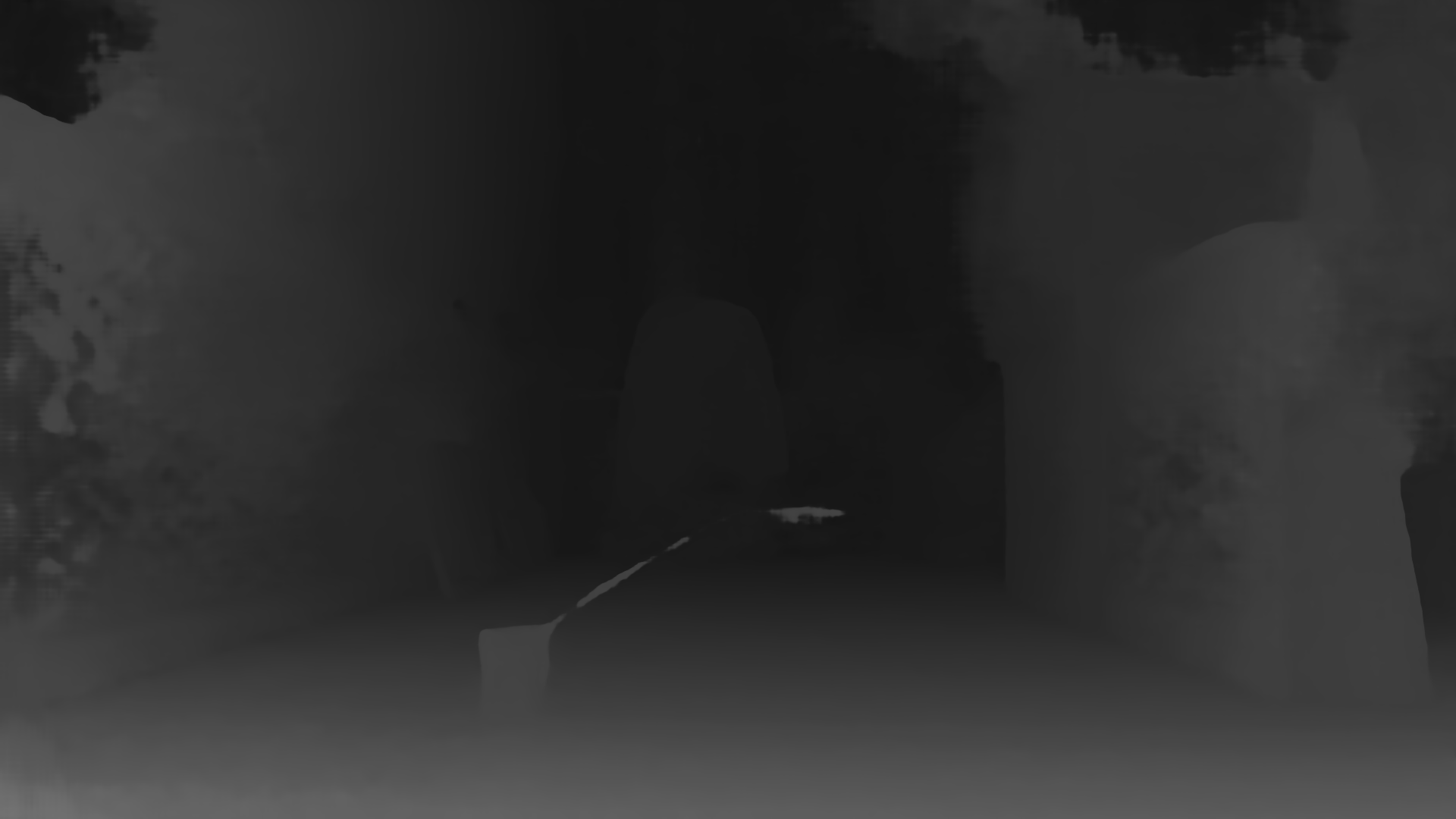}}
\subfigure[RAftNet]{\includegraphics[width=0.23\textwidth]{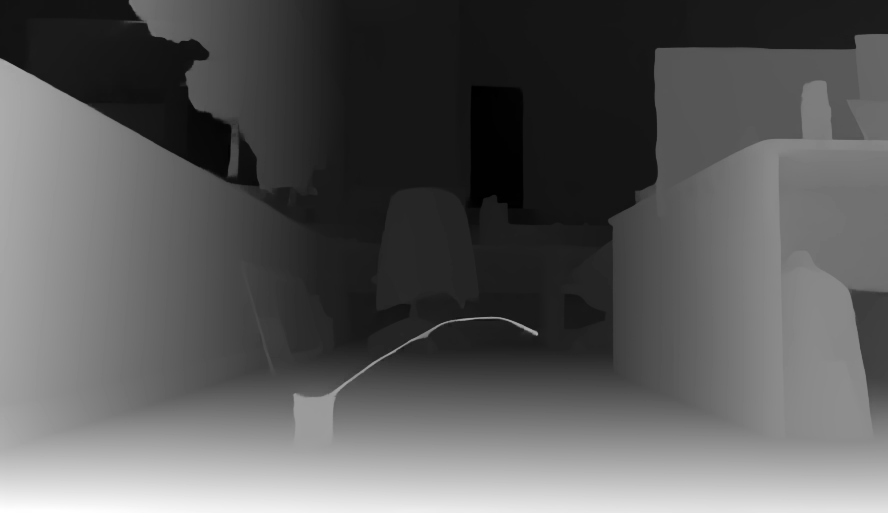}}
\subfigure[Nerf stereo]{\includegraphics[width=0.23\textwidth]{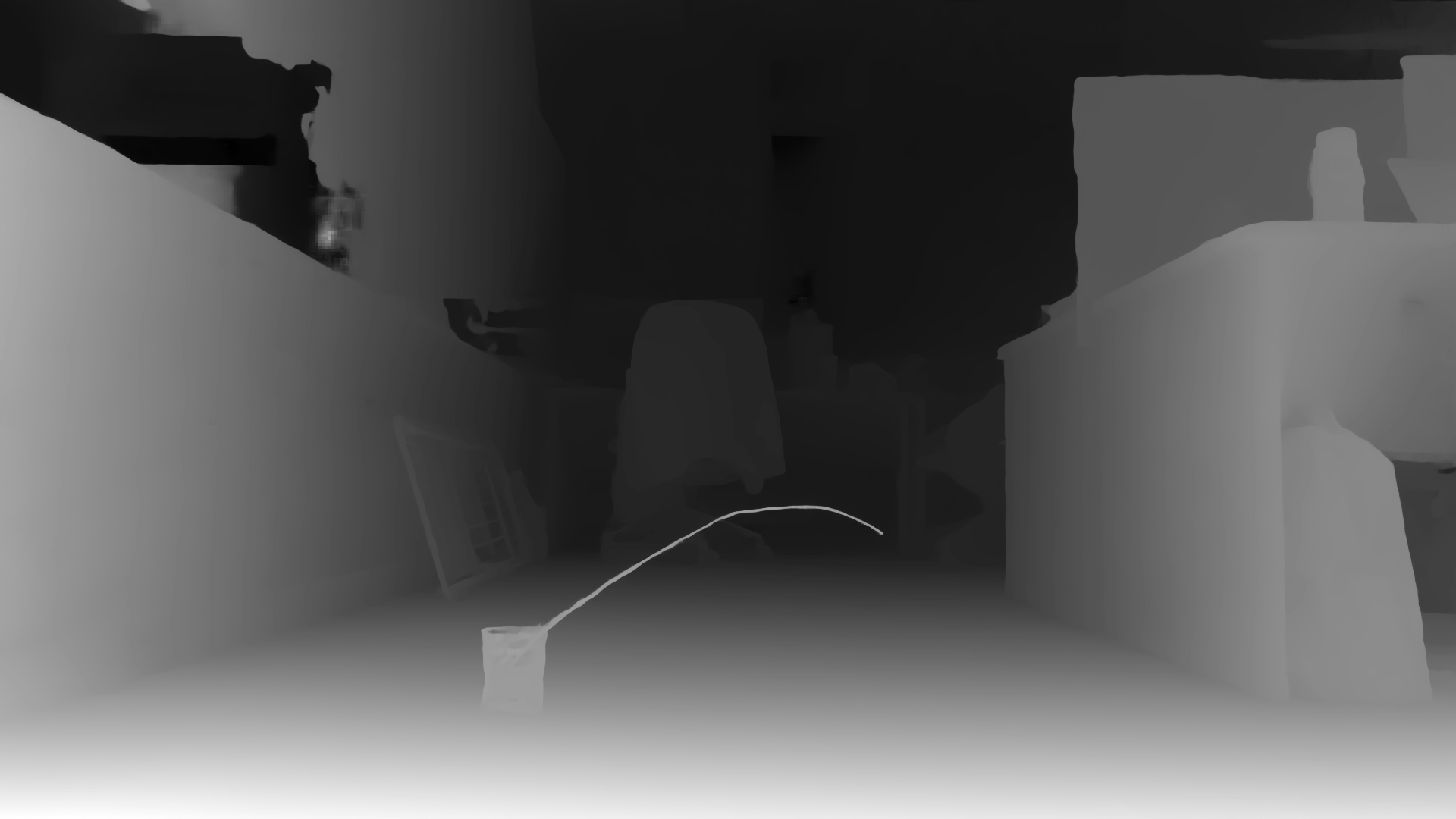}}
\caption{Comparison of Depth Map Results from Different Stereo Matching Models}
\label{stereo}
\end{figure}

%stereo
Then, we employed fine-tuning on various stereo neural network methods to predict our dataset, with the results presented in Fig. \ref{stereo}. The NeRF method achieved the highest performance, albeit with the longest processing time, approximately 6 seconds.

\subsection{Final Results Achieved by Combining YOLO with NeRF Using the Second Combination Method}

\begin{figure}[htbp]
\centering
\subfigure[Depth Map Generated by SGBM at 1m Distance from Camera to Branch]{\includegraphics[width=0.23\textwidth]{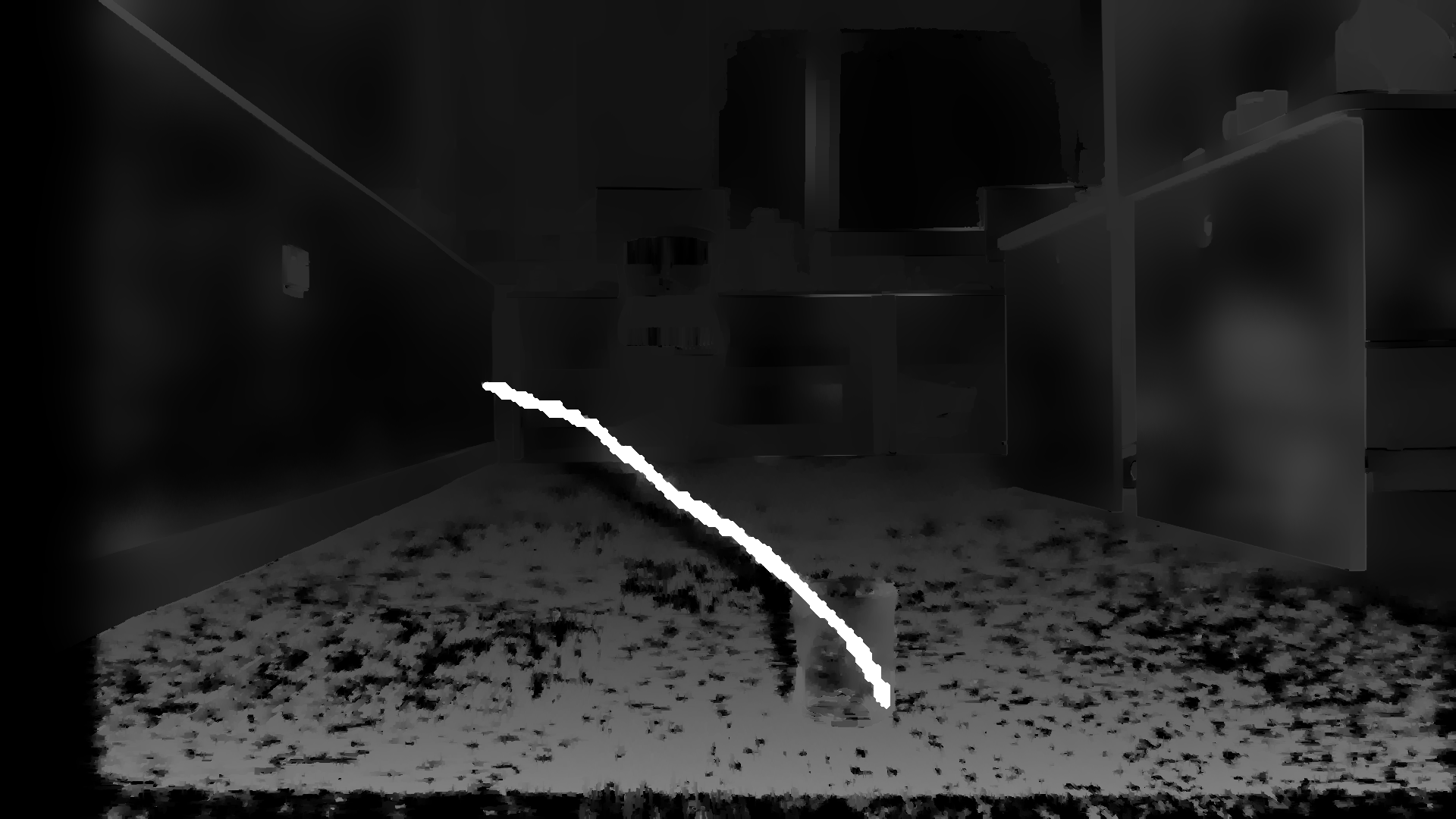}}
\subfigure[Depth Map Generated by SGBM at 1.5m Distance from Camera to Branch]{\includegraphics[width=0.23\textwidth]{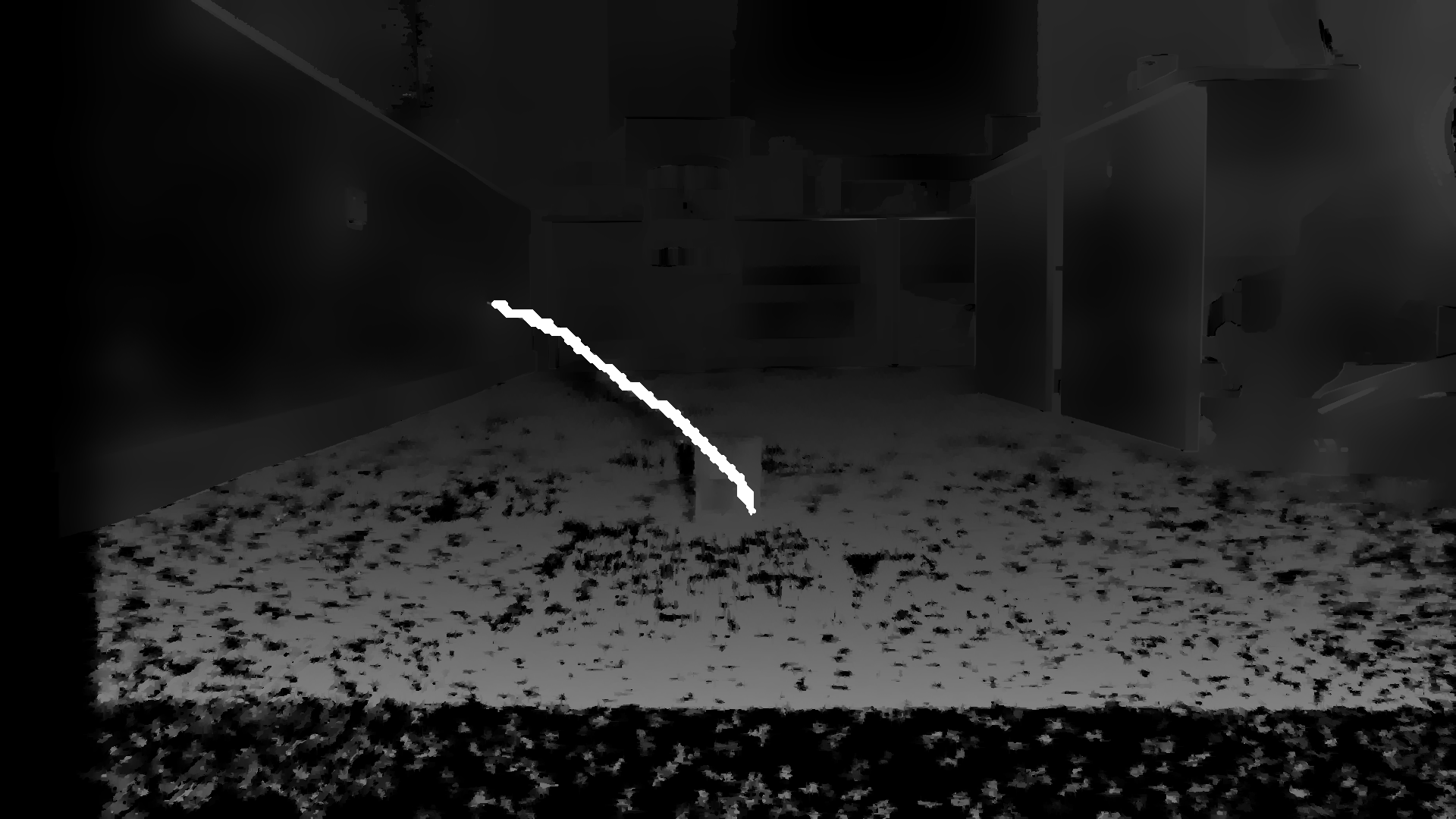}}
\subfigure[Depth Map Generated by SGBM at 2m Distance from Camera to Branch]{\includegraphics[width=0.23\textwidth]{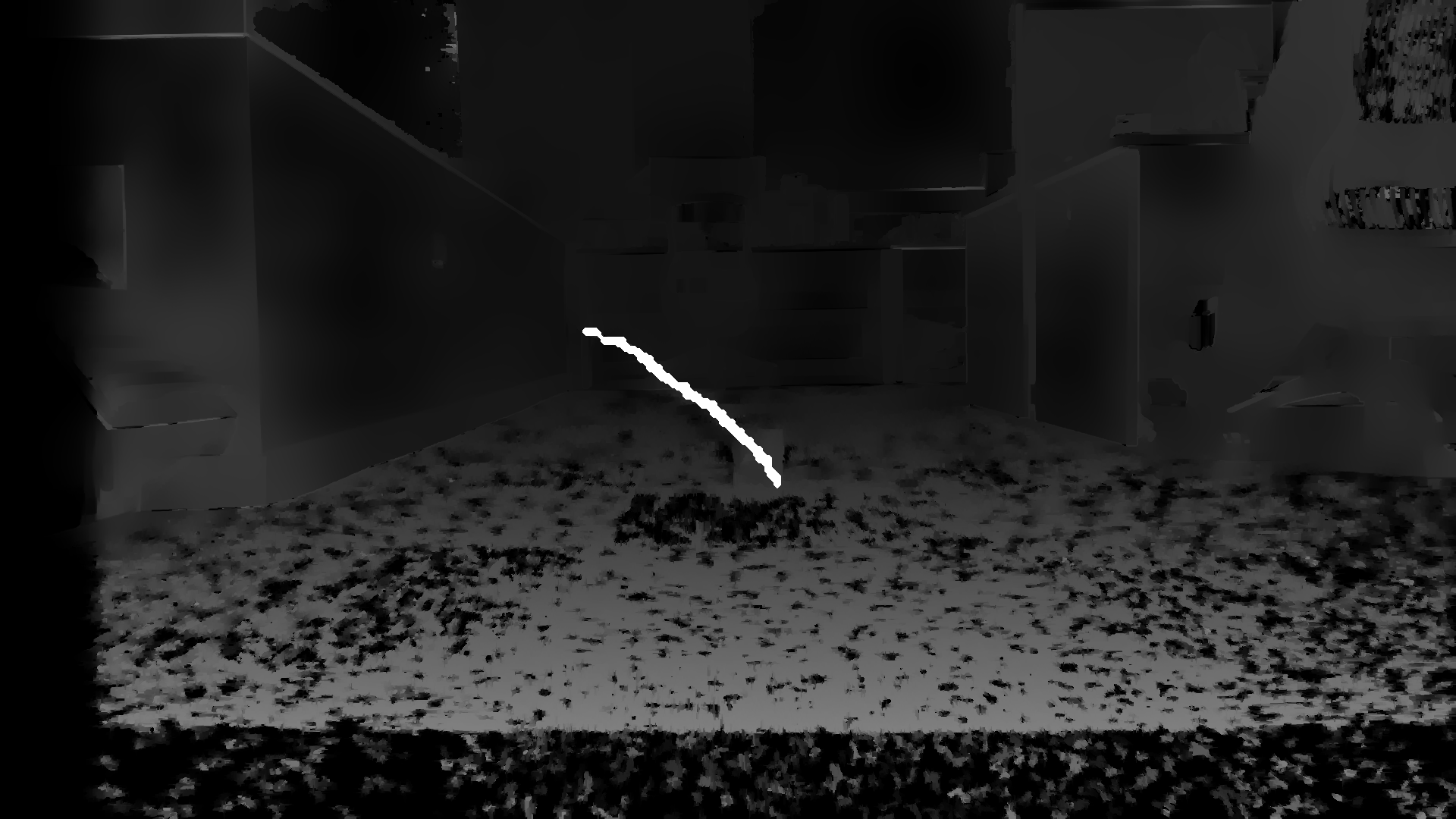}}
\subfigure[Depth Map Generated by Nerf at 1m Distance from Camera to Branch]{\includegraphics[width=0.23\textwidth]{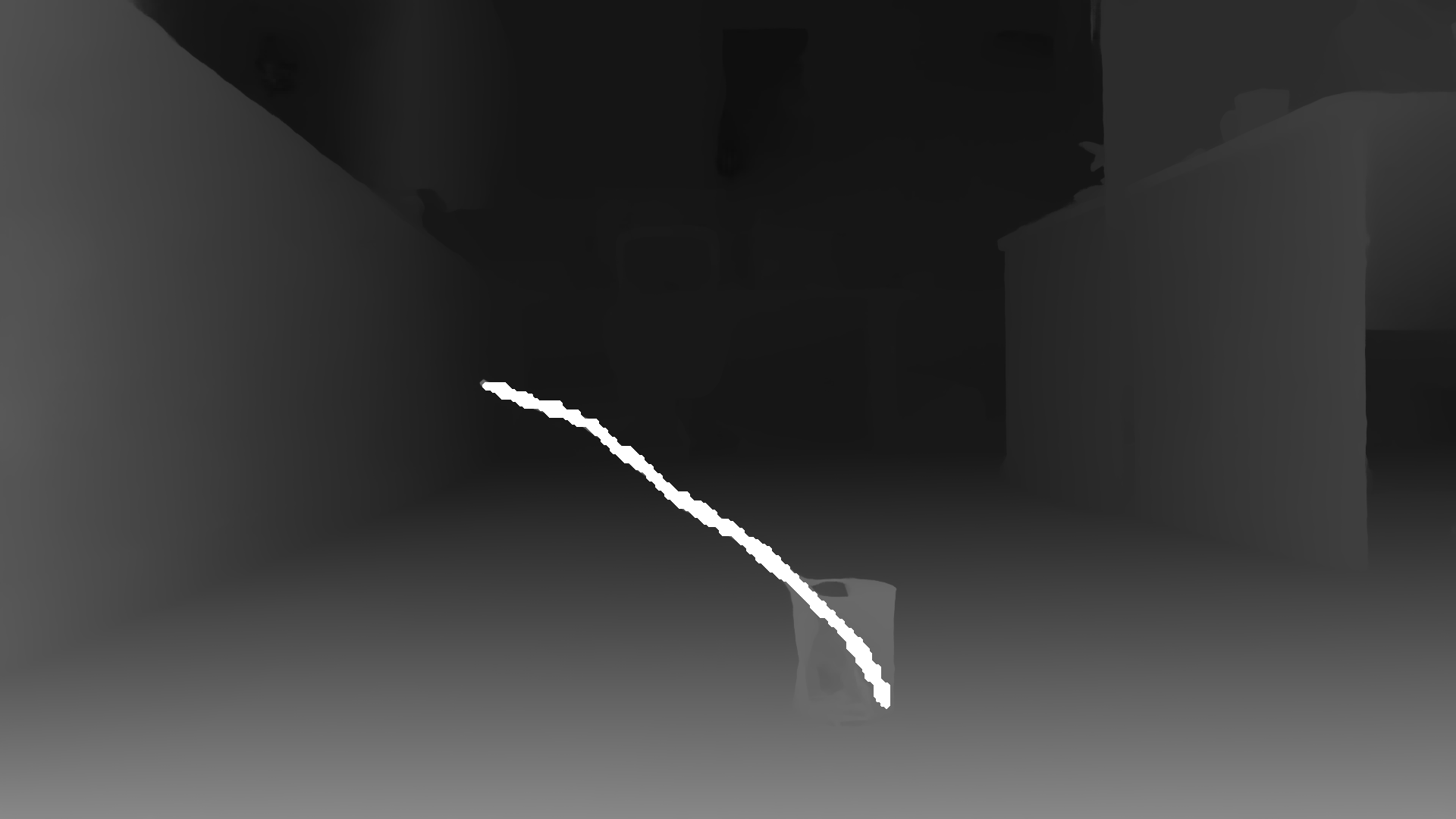}}
\subfigure[Depth Map Generated by Nerf at 1.5m Distance from Camera to Branch]{\includegraphics[width=0.23\textwidth]{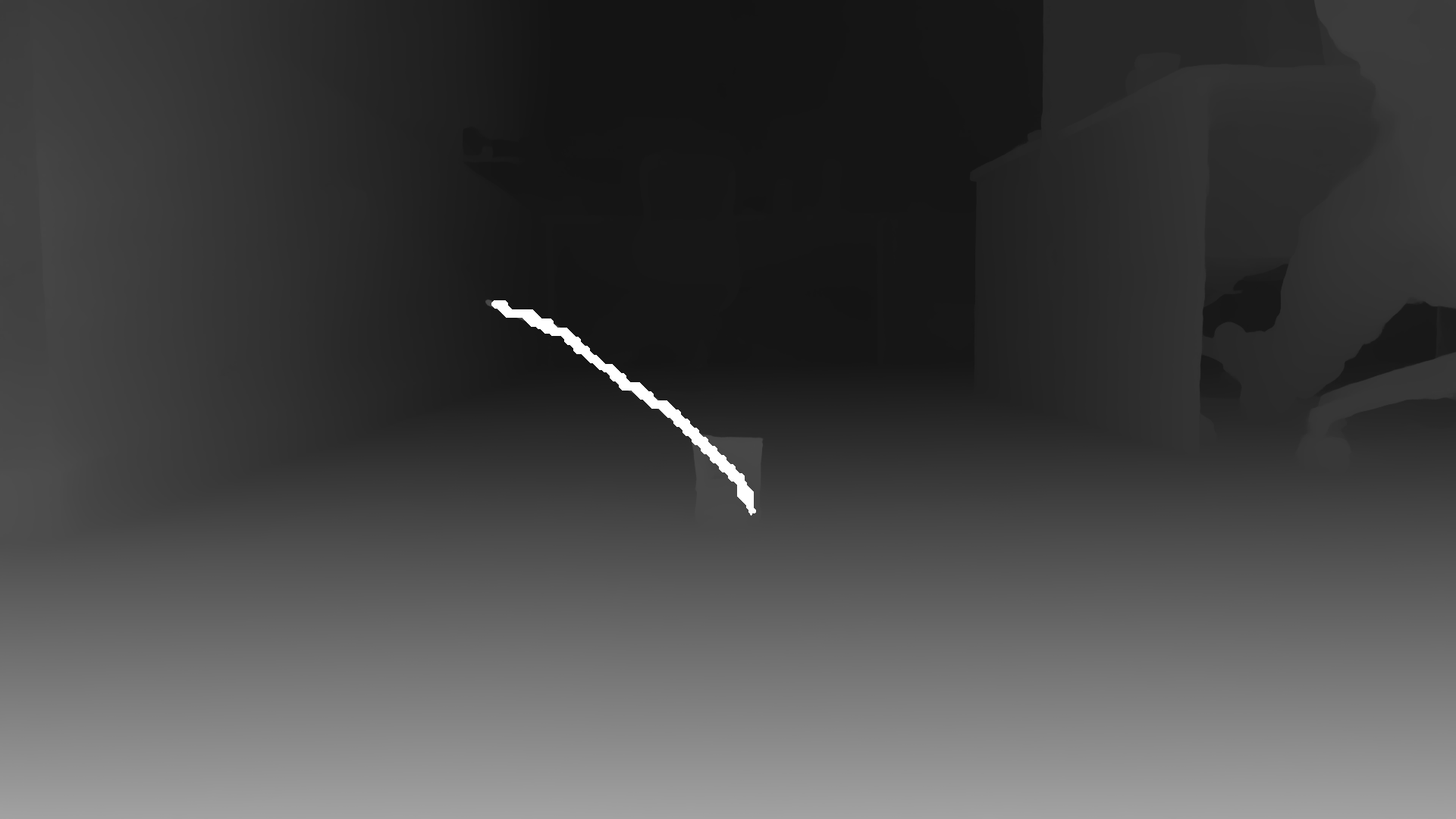}}
\subfigure[Depth Map Generated by Nerf at 2m Distance from Camera to Branch]{\includegraphics[width=0.23\textwidth]{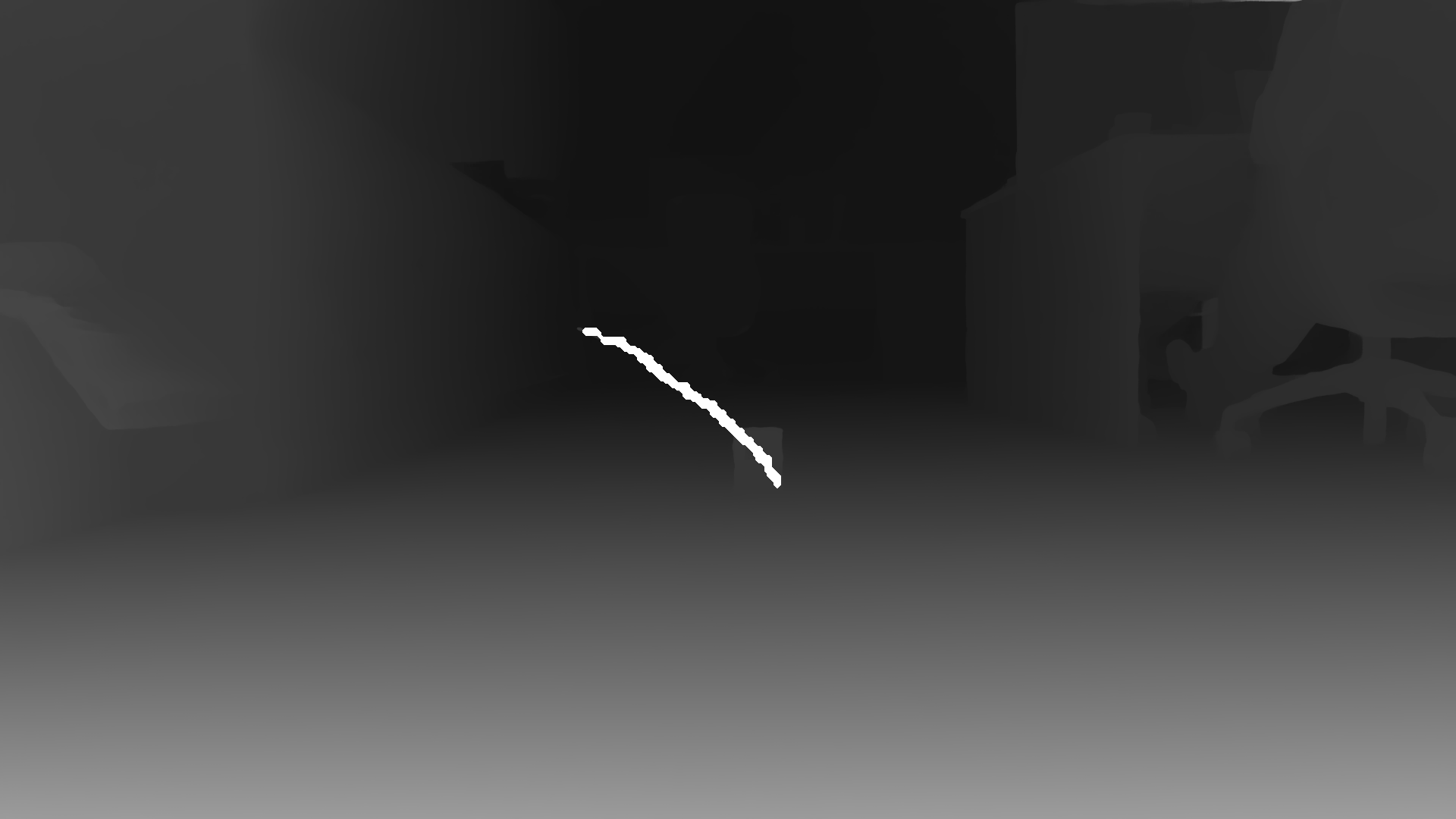}}
\caption{Comparison of YOLO Combined with SGBM and NeRF Across Varying Distances: (a)-(c) Representing SGBM at 1m, 1.5m, and 2m; (d)-(f) Representing NeRF at 1m, 1.5m, and 2m}
\label{SGBM_and_Nerf}
\end{figure}

\begin{figure}[htbp]
\centering
\subfigure[Histogram of SGBM Generated Depth Map at 1m Distance Between Branch and Camera]{\includegraphics[width=0.23\textwidth]{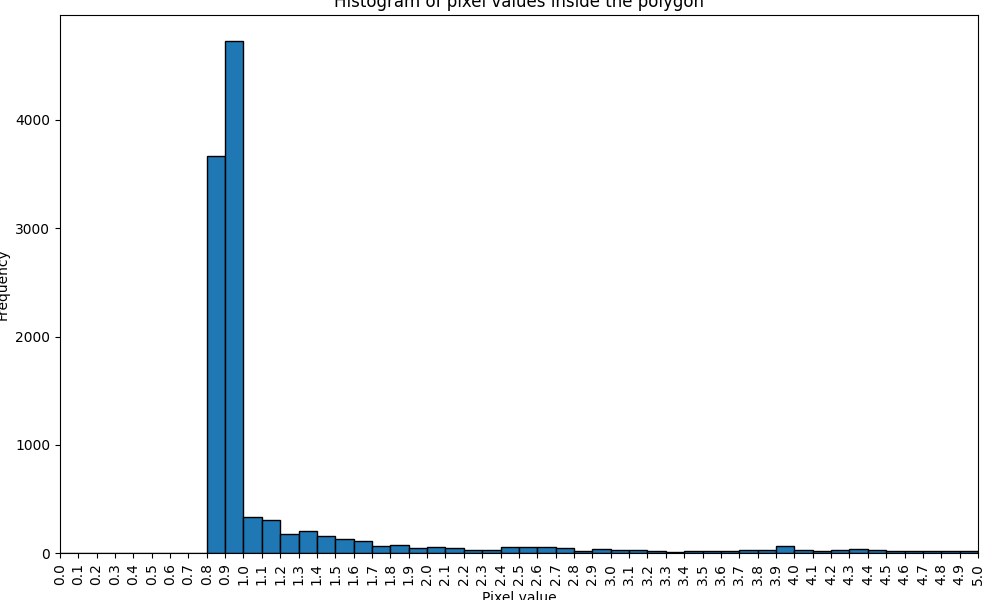}}
\subfigure[Histogram of SGBM Generated Depth Map at 1.5m Distance Between Branch and Camera]{\includegraphics[width=0.23\textwidth]{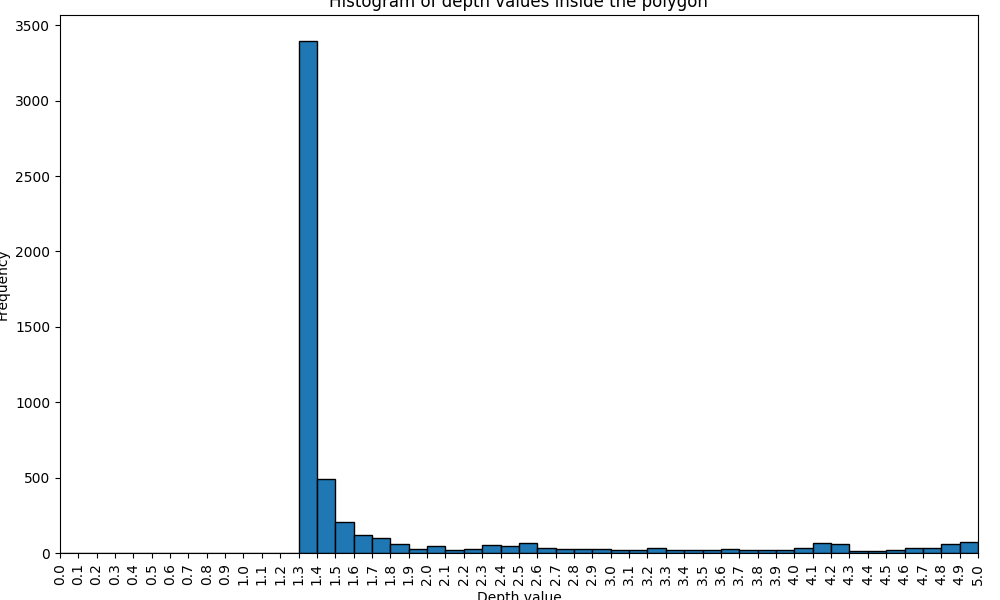}}
\subfigure[Histogram of SGBM Generated Depth Map at 2m Distance Between Branch and Camera]{\includegraphics[width=0.23\textwidth]{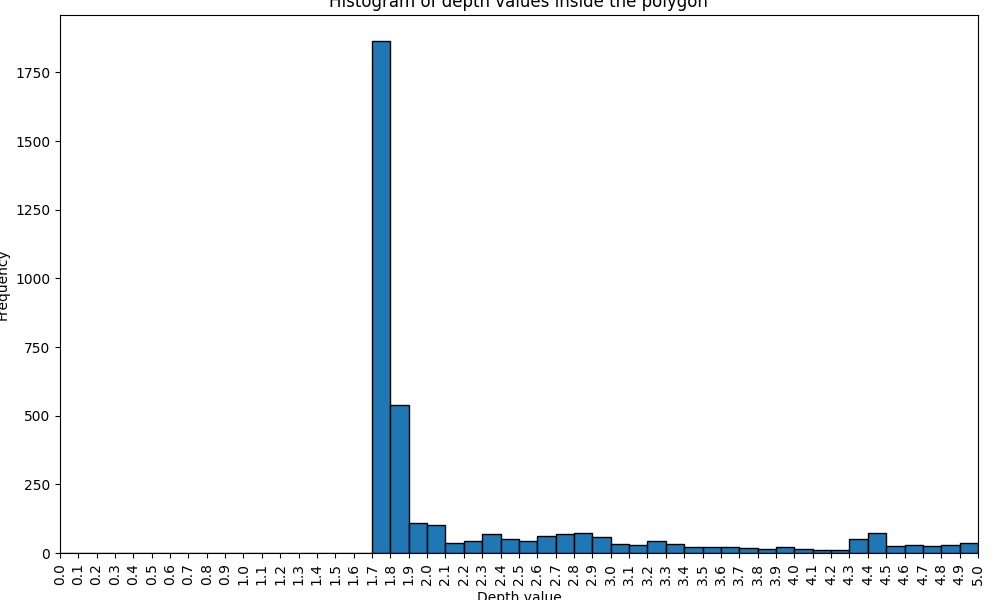}}
\subfigure[Histogram of Nerf Generated Depth Map at 1m Distance Between Branch and Camera]{\includegraphics[width=0.23\textwidth]{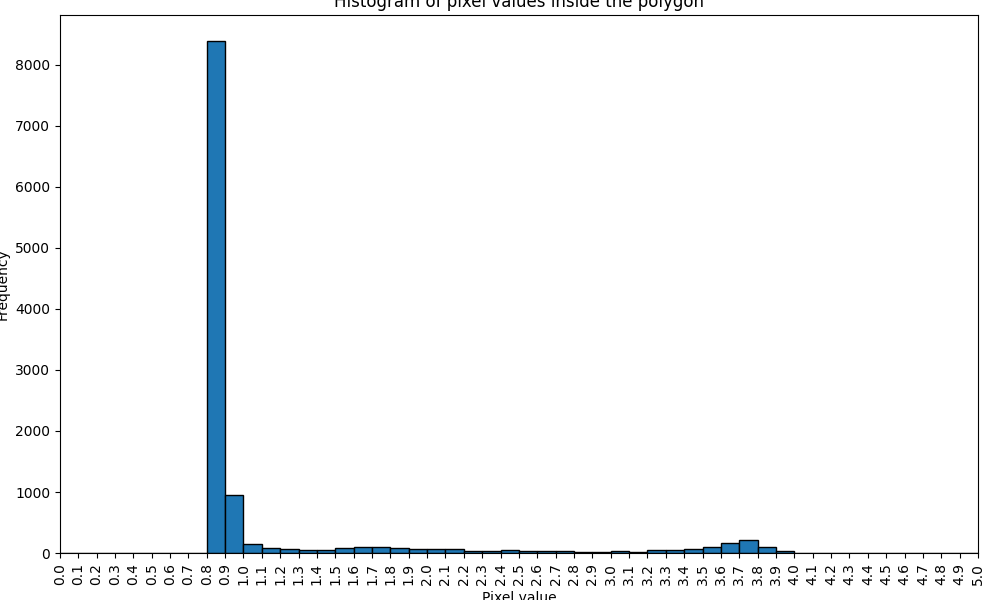}}
\subfigure[Histogram of Nerf Generated Depth Map at 1.5m Distance Between Branch and Camera]{\includegraphics[width=0.23\textwidth]{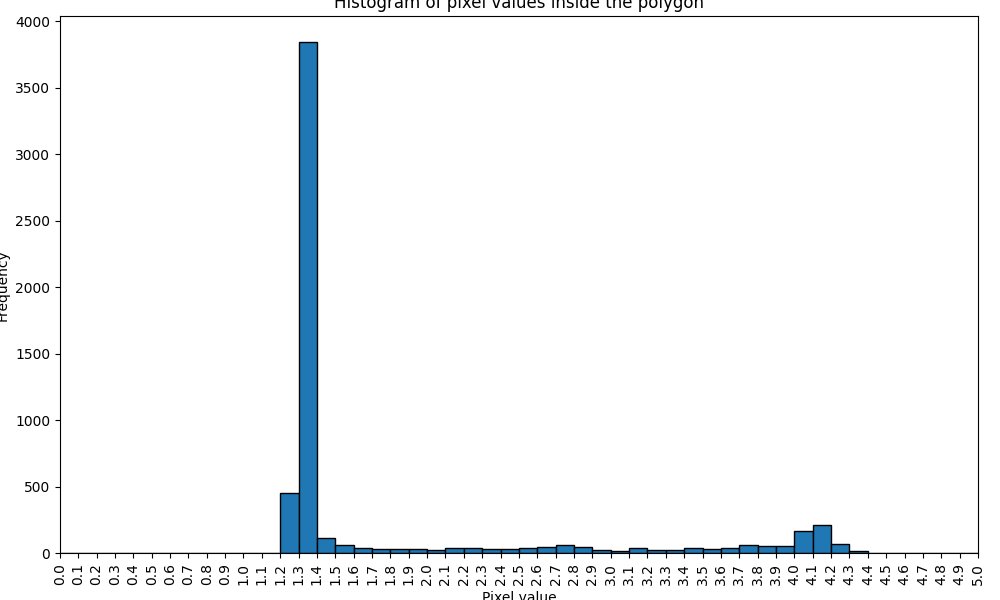}}
\subfigure[Histogram of Nerf Generated Depth Map at 2m Distance Between Branch and Camera]{\includegraphics[width=0.23\textwidth]{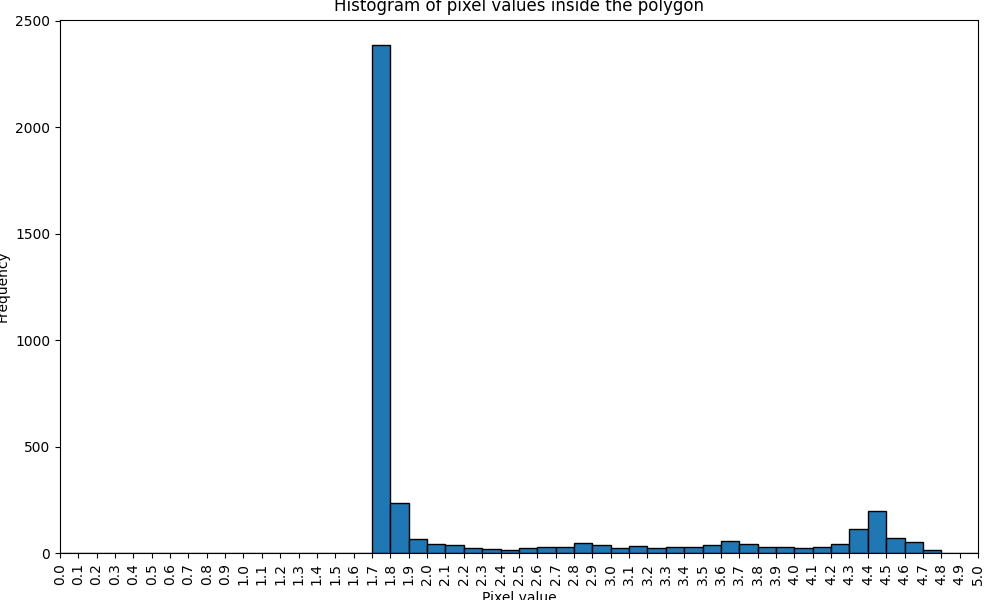}}
\caption{Distribution Plots of YOLO Combined with SGBM and NeRF Across Varying Distances. with(a)-(c) Representing SGBM at 1m, 1.5m, and 2m, and (d)-(f) Representing NeRF at 1m, 1.5m, and 2m.}
\label{distri_SGBM_and_Nerf}
\end{figure}

We employed YOLOv8n-seg as the instance segmentation model. Utilizing the original combination method\cite{linDroneStereoVision2024}, a comparative analysis was conducted between two leading depth estimation models, SGBM and NeRF, as demonstrated in Fig. \ref{SGBM_and_Nerf}. The resulting data distribution is presented in Fig. \ref{distri_SGBM_and_Nerf}. The analysis indicates that while SGBM primarily detects points within a more accurate depth range, it yields fewer detected points compared to NeRF. Consequently, future research will focus on enhancing depth map accuracy for tree branches through the application of NeRF, with the aim of approximating ground-truth depth data. Nonetheless, a key limitation of NeRF is its processing time, currently requiring approximately 6 seconds to generate a depth map. Addressing this computational efficiency will be a major priority in our subsequent investigations.

\begin{figure}[htbp]
\centering
\subfigure[Final Centroid Points Calculated from Three Predicted Points]{\includegraphics[width=0.23\textwidth]{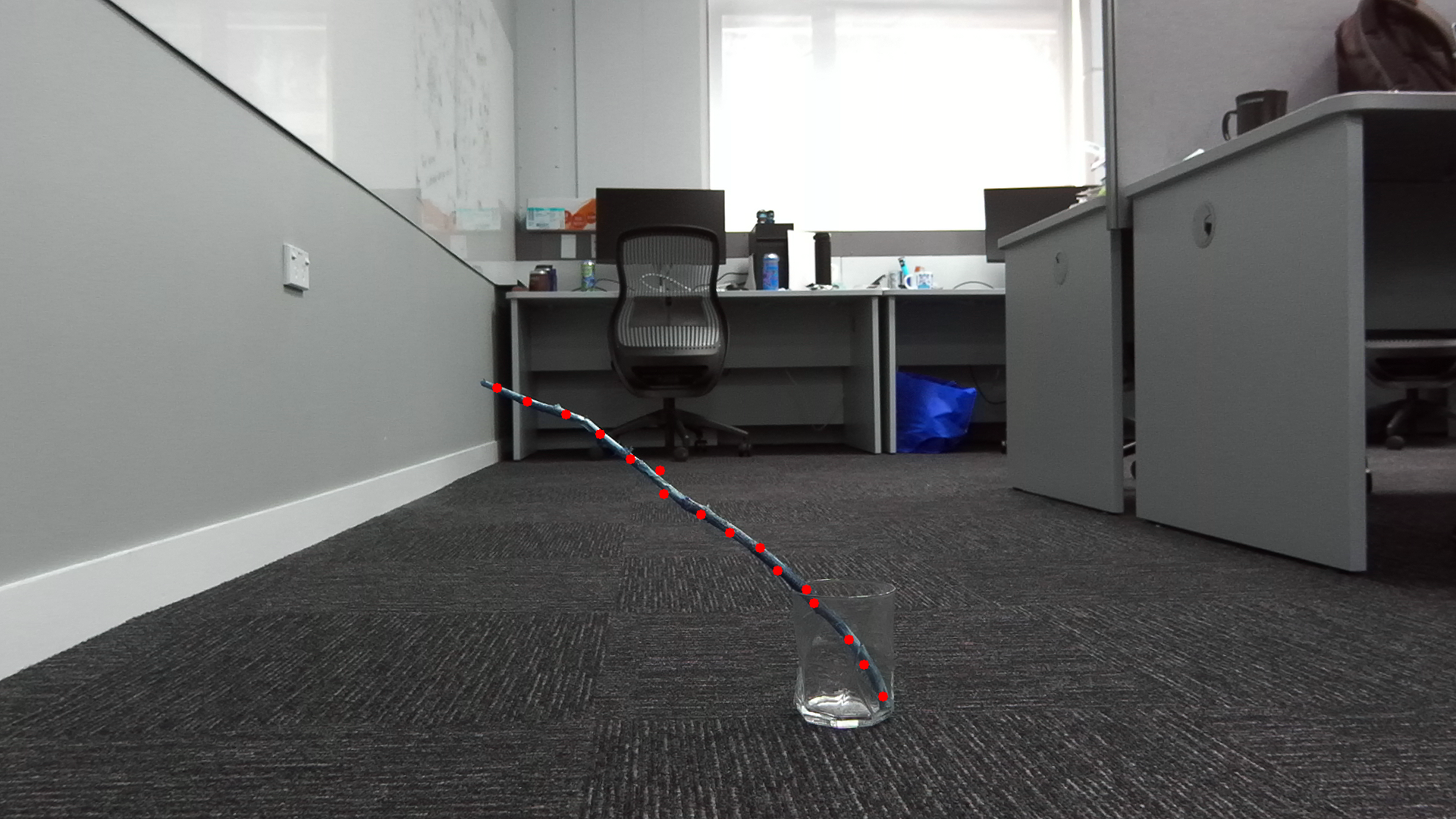}}
\subfigure[Depth Map Generated Using NeRF]{\includegraphics[width=0.23\textwidth]{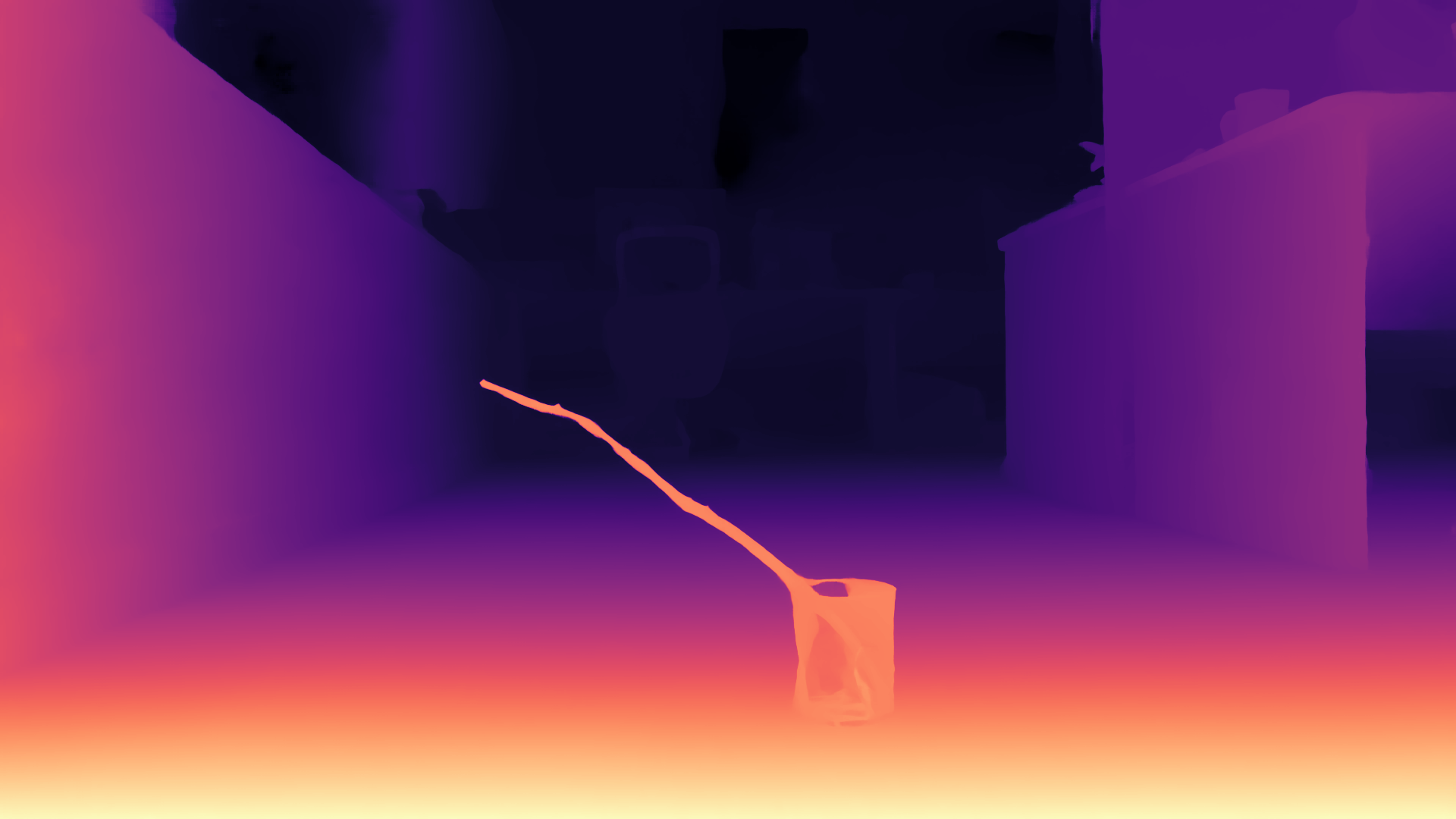}}
\caption{(a) Present the results of the second method, where the final centroid points are calculated from a set of three predicted points spaced a certain distance apart, and (b) display the depth map obtained from NeRF. Combining these allows for determining the final distance between the branches and the stereo camera.}
\label{circle_image}
\end{figure}

In Fig. \ref{circle_image}, we present the triangular integration method developed in this research. This method optimizes memory usage by capturing only a select number of key points, rather than mapping all points within the plane, to accurately record the positions of tree branches. Such an approach is expected to be highly beneficial in future applications, particularly in scenarios where drones are required to navigate and operate in complex environments with dense tree branch structures.

\section{Conclusions}

In this research, we have integrated advanced branch detection and segmentation techniques with high-precision depth maps generated by deep learning models to enhance the accuracy of distance measurements between tree branches and drones, thereby facilitating precise and efficient pruning operations. While depth map generation through deep learning methods such as NeRF has demonstrated exceptional accuracy, its slow processing speed remains a significant limitation. Our integration aims to mitigate this issue by leveraging the strengths of both advanced detection techniques and deep learning models. Recognizing the constraints imposed by our current small dataset, future work will involve utilizing drones to collect a larger and more diverse dataset with increased environmental complexity. This expanded dataset is expected to enhance the robustness of our models and provide a valuable foundation for continued research in this domain.

\bibliographystyle{IEEEtran} % 引用樣式文件
%\bibliography{references} % 引用的文獻文件
% Generated by IEEEtran.bst, version: 1.14 (2015/08/26)

\vspace{12pt}

\end{document}